\documentclass[11pt]{article}

\usepackage[final]{acl}

\usepackage{times}
\usepackage{latexsym}

\usepackage{amsmath,amssymb}
\usepackage{booktabs}
\usepackage{multirow}
\usepackage{makecell}
\usepackage{rotating}

\newcommand{\score}[2]{\ensuremath{#1_{\pm #2}}}
\newcommand{\bestscore}[2]{\ensuremath{\mathbf{#1}_{\pm #2}}}

\usepackage[T1]{fontenc}

\usepackage[utf8]{inputenc}

\usepackage{microtype}

\usepackage{inconsolata}

\usepackage{graphicx}

\title{Router Prior Bias: Preserving Base Routing Structure in MoE Post-Training}

\author{
  \textbf{Jaedeok Lee\textsuperscript{1,2}},
  \textbf{Keonwoo Kim\textsuperscript{1}},
  \textbf{Dongyoon Han\textsuperscript{3}},
\\
  \textbf{Sangdoo Yun\textsuperscript{3}},
  \textbf{Yera Choi\textsuperscript{1,\dag}},
  \textbf{Haanju Yoo\textsuperscript{1,\dag}}
\\[3pt]
  \textsuperscript{1}NAVER Applied AI Group
\\
  \textsuperscript{2}Healthcare AI Research Institute (HARI),
  Seoul National University Hospital
\\
  \textsuperscript{3}NAVER AI Lab
\\[3pt]
  \small{
    \textsuperscript{\dag}Corresponding authors.
    \quad
    \textbf{Correspondence:} \href{mailto:yera.choi@navercorp.com}{yera.choi@navercorp.com}, \href{mailto:haanju.yoo@navercorp.com}{haanju.yoo@navercorp.com}
  }
}

\begin{document}
\maketitle
\begin{abstract}
Mixture-of-Experts (MoE) pretraining relies on an auxiliary load-balancing loss (LBL) to drive per-expert utilization toward uniformity.
Post-training inherits a different situation: the base router already encodes non-uniform expert co-activation structure, which a re-imposed uniformity objective flattens away.
We show that downstream performance depends instead on holding this inherited routing \emph{softly}, a principle we term soft router anchoring, and instantiate it as \textbf{Router Prior Bias (RPB)}, a training-time bias that pulls the router logits toward a prior read off the frozen base router while leaving the router itself trainable.
On math post-training of Moonlight-16B-A3B, RPB attains 45.77 in-domain accuracy against 31.91 under re-applied LBL and 29.44 under unanchored fine-tuning, and retains more out-of-domain capability than either.
The ordering against LBL reproduces on a second model family (Qwen3-30B-A3B-Base), and the advantage over LBL is resolvable on an independently sourced corpus.
Anchors defined on the router weights, on its logits, or on its output distribution perform comparably with no consistent ordering, which places the effect in the softness of the constraint rather than in the particular prior RPB supplies.
Retained community structure in the expert co-activation graph tracks these gains wherever the base router is non-uniform enough for communities to form, yet enforcing the same prior as a hard assignment preserves that structure while performance falls sharply.
Community structure is therefore a footprint of soft anchoring rather than its source, and the practical lesson is that inherited routing should be held softly during post-training, since both flattening it toward uniformity and enforcing it absolutely carry a downstream cost.
Our code will be released at \href{https://github.com/naver-ai/rpb}{github.com/naver-ai/rpb}.
\end{abstract}

\section{Introduction}

Mixture-of-Experts (MoE) language models, including Qwen~\citep{qwen3technicalreport}, Mixtral~\citep{jiang2024mixtralexperts}, DeepSeek~\citep{deepseekv2}, and Moonlight~\citep{liu2025muonscalablellmtraining}, route each token to a small subset of experts, allowing total parameter count to grow without a commensurate increase in per-token computation~\citep{Cai2024ASO}.
Because an unbalanced router would leave much of that capacity idle, MoE pretraining adds an auxiliary load-balancing loss (LBL) that penalizes deviations from uniform per-expert utilization~\citep{Lepikhin2020GShardSG,Fedus2021SwitchTS}.

By the time post-training begins, however, the router is no longer an unstructured allocator: pretraining has already induced non-uniform specialization and recurring patterns of expert co-activation~\citep{Tang2026SpecializationTC,Lo2024ACL}, so a re-imposed uniformity objective works against structure the model has spent its pretraining budget acquiring.
Academic adaptation methods and practitioner toolchains have accordingly converged on preserving that inherited structure rather than re-imposing pretraining-style uniformity: structure-aware adaptation methods select, retain, or route through it~\citep{wang2024let,Li2025DynamicES,eo2025mixture}, while post-training toolchains drop load balancing altogether, with NVIDIA Megatron-Bridge exposing no-load-balancing and auxiliary-loss-free configurations~\citep{nemo-megatron-bridge} and Unsloth disabling router-layer training entirely~\citep{unsloth-moe}.
These defaults are widely adopted, yet which property of the base router they preserve, and why that preservation benefits adaptation, remain underexamined.

One reason the question has stayed open is that the available defaults sit at the two extremes, either leaving base routing unconstrained or enforcing it in full.
Router-freeze, the strictest of them, obtains routing stability by removing trainability altogether, which caps how far the model can adapt to the post-training corpus and answers two distinct questions at once: which routing structure to preserve, and how strictly to enforce it.
Separating the two requires an intervention whose enforcement strength is a continuous quantity rather than an architectural commitment.
We therefore introduce \textbf{Router Prior Bias (RPB)}, a training-time soft logit bias that anchors routing to a prior formed from the frozen base router while leaving the router itself trainable.

\begin{figure*}[t]
\centering
\includegraphics[width=\textwidth]{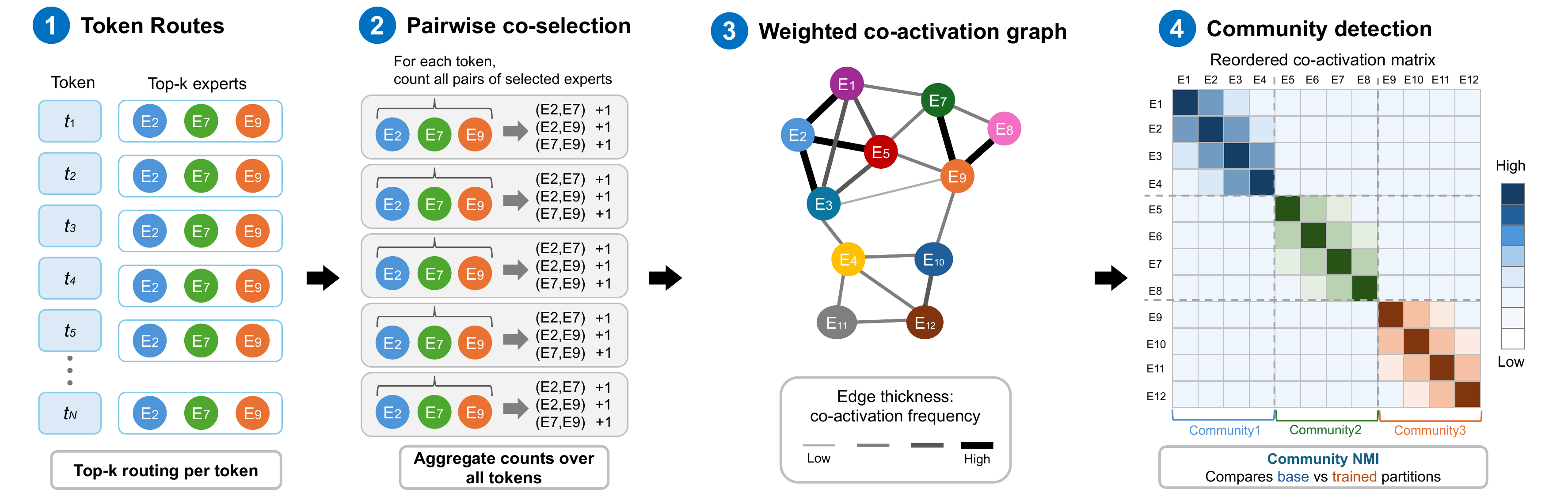}
\caption{
  Construction of the layer-wise expert co-activation graph, in four steps. For each input token, the router selects its top-$k$ experts, where $E_i$ denotes expert $i$. Experts selected together for the same token are counted as co-activations. These co-activation counts are accumulated into a layer-wise weighted expert graph, with experts as nodes and co-activation frequency as edge weight. Community detection then partitions the graph into expert communities, and community NMI compares the base and post-trained partitions.}
\label{graph}
\end{figure*}

To identify what such anchoring actually preserves, we measure routing retention at three levels: (1) \emph{per-expert utilization}, the level classical LBL targets; (2) \emph{token-level top-$k$ selection}, which experts a given token is routed through; and (3) \emph{expert co-activation community structure}, which experts tend to be selected together (Figure~\ref{graph}).
Methods in the \emph{soft router anchoring} family, RPB among them, reach high downstream performance while retaining base top-$k$ selection and community structure, whereas re-applied LBL preserves per-expert utilization alone.
Objectives that constrain the router parameters directly, without constructing any prior at all, behave the same way, so what the family shares is the anchoring rather than the prior RPB supplies.

This leaves open whether the gain comes from soft enforcement itself or from the community structure that anchoring happens to preserve, and we separate the two interventionally.
Enforcing the same prior as a hard assignment preserves community structure at or above the level soft anchoring reaches, yet substantially reduces performance (Appendix~\ref{sec:appendix-beta-operating}).
Shuffling the prior then varies its content at fixed enforcement: a within-community shuffle applied softly leaves performance essentially intact, while the same prior applied as a hard constraint does not (\S\ref{sec:results-shuffled}).
Preserved community structure is therefore a footprint of soft enforcement at training time rather than the source of the performance gain.

Our contributions are as follows:
\begin{enumerate}\itemsep2pt
\item \textbf{A controlled comparison of post-training routing objectives.} Re-applying the pretraining load-balancing loss during MoE post-training degrades downstream performance relative to leaving the inherited routing softly anchored, and the effect persists under a change of model family (Qwen3-30B-A3B-Base) and of corpus (OpenR1-Math-220k).
Whether soft anchoring additionally outperforms an unanchored no-load-balancing baseline is model-dependent, and we report both directions.
\item \textbf{Soft router anchoring, and a method that implements it.} On Moonlight-16B-A3B, four objectives acting in parameter, logit, and probability space perform comparably with no consistent ordering, which locates the effect in the anchoring itself rather than in any one objective and makes it available to any method that holds the router near its base routing while leaving it trainable. RPB is the sample-conditional instance we develop and evaluate.
\item \textbf{A retention protocol and a diagnostic.} Measuring retention at three levels, with the training-time prior removed at evaluation, identifies community NMI as the level that separates soft anchoring from re-applied LBL, \emph{in base models whose routing is non-uniform enough to form communities}, a condition Moonlight-16B-A3B satisfies and DeepSeek-V2-Lite does not.
\item \textbf{Cause separated from footprint.} Interventional controls that vary enforcement strength and prior content independently isolate soft enforcement, rather than community-structure preservation, as the source of the performance gain.
\end{enumerate}

\section{Preliminaries}
\label{sec:prelim}
\label{sec:method-graph}

We now make precise the three levels at which we measure \emph{routing retention}, the degree to which a post-trained router still routes as its base router did: (1) per-expert utilization, (2) token-level top-$k$ selection, and (3) expert co-activation community structure, all reported alongside downstream \textbf{performance} (in-domain and out-of-domain means).
Per-expert utilization is a sanity condition rather than a result, since every method we compare falls in the same range on it.
All three levels are measured \emph{bias-free}\label{sec:setup-biasfree}: the training-time RPB prior-bias hook is removed at evaluation, so the community structure and top-$k$ overlap we measure reflect the trained router rather than an inference-time prior.

\paragraph{Per-expert utilization.}
This is the level classical LBL targets.
Between the post-training and base routers' expert distributions we report total variation distance (TVD) for $L_1$ marginal shift and Jensen--Shannon divergence (JSD) for symmetric, bounded shift.

\paragraph{Token-level top-$k$ selection.}
For per-token selection identity we report \textbf{top-$k$ overlap}: for each token, the intersection size between the base-router and post-training top-$k$ sets, normalized by $k$ and averaged over tokens and layers.

\paragraph{Expert co-activation community structure.}
For each transformer layer $\ell$, we build an expert co-activation graph (Figure~\ref{graph}) whose nodes are experts and whose edge weight $w_{ij}^{(\ell)}$ records how often experts $i$ and $j$ appear together in the same token's top-$k$ route, averaged over probe tokens (self-loops excluded so the metric measures inter-expert collaboration):
\begin{equation}
w_{ij}^{(\ell)}
= \frac{1}{N}\sum_{t=1}^{N}
\mathbf{1}\bigl\{i, j \in \mathrm{top}_k(x_t)\bigr\},
\quad i \neq j,
\end{equation}
where $N$ is the number of probe tokens, drawn from a held-out probing pool of roughly $5{,}000$ tokens per domain (Appendix~\ref{app:probe-size}), and $\mathrm{top}_k(x_t)$ is the router's top-$k$ set for token $x_t$ at layer $\ell$.
We threshold edges by a retained-edge fraction $\rho$ (default $0.10$, with a robustness analysis in Appendix~\ref{app:sweeps}) and run Louvain community detection~\cite{Blondel2008FastUO} layer-wise, partitioning experts to maximize modularity.
For partition alignment we report \textbf{community NMI}, the layer-averaged normalized mutual information (NMI) between the post-training and base partitions on the same probe corpus.
We also report modularity $Q$~\cite{Newman2006ModularityAC} for absolute clustering strength and $\Delta Q$ for its change relative to the base model.

\section{Method}

\subsection{Router Prior Bias (RPB)}
\label{sec:method-rpb}

\begin{figure*}[t]
\centering
\includegraphics[width=\textwidth]{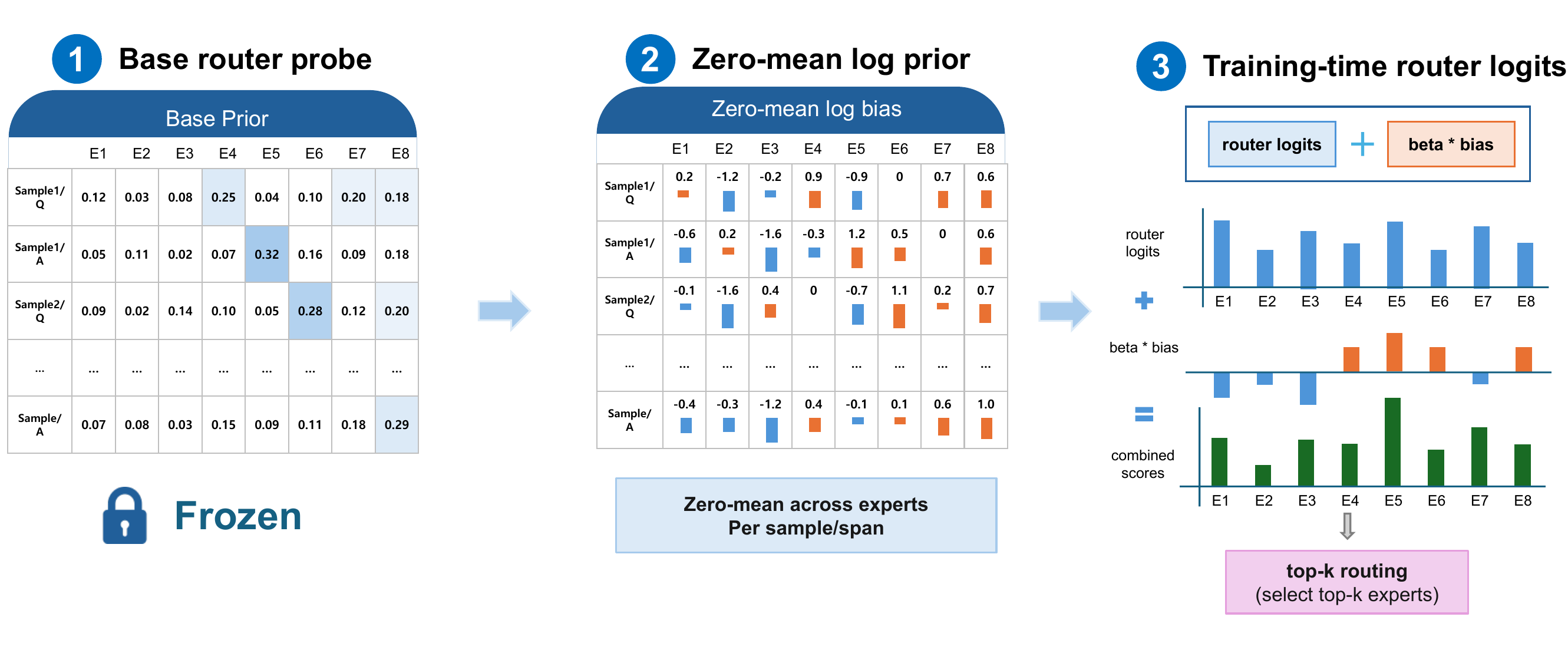}
\caption{RPB first estimates a frozen-base expert prior for each sample/span and layer, then
  converts it into a zero-mean log-prior bias across experts. During training, this
  bias is added to the current router logits with strength $\beta$ before top-$k$ selection,
  softly anchoring routing toward the base model. The bias is removed at evaluation, so the
  measured routing reflects the trained router alone.}
\label{RPB}
\end{figure*}

RPB is a soft, training-time logit bias derived from a frozen base router and added to the current router's logits during post-training (Figure~\ref{RPB}).

\paragraph{Frozen-base prior over Q/A spans.}
For each training sample $x$, span $s \in \{Q, A\}$, layer $\ell$, and expert $e$, we precompute the frozen-base routing prior
\begin{equation}
\pi_{x,s,\ell,e}
= \mathrm{mean}_{t \in (x,s)}\,
p_{\mathrm{base}}(e \mid x_t,\ell),
\end{equation}
where $p_{\mathrm{base}}$ is the frozen base router's softmax gate.
A \emph{span} is one of two semantic regions of the training example: \textbf{Q} (instruction/prompt) and \textbf{A} (response/answer).
We use Q/A spans because Q and A induce different base routing distributions (Appendix~\ref{sec:appendix-probe}), and aggregating them would conflate input-domain with output-format routing.

\paragraph{Training-time bias.}
Let $c_{x,s,\ell,e} = \mathrm{clip}(\pi_{x,s,\ell,e})$ and let $z_{t,\ell}$ be the router-logit vector across experts at layer $\ell$ for token $t$.
RPB adds a zero-mean log-prior bias to the logits:
\begin{align}
b_{x,s,\ell,e}
&= \log c_{x,s,\ell,e}
 - \mathrm{mean}_{e'} \log c_{x,s,\ell,e'}, \\
z'_{t,\ell}
&= z_{t,\ell} + \beta\, b_{x,s,\ell},
\quad t \in (x,s).
\end{align}

Here $\mathrm{clip}(p)$ clamps probabilities to $[\epsilon/E, C/E]$, where $E$ is the number of experts in the layer (defaults $\epsilon{=}0.05$, $C{=}5.0$), keeping the log-bias finite, and the expert-dimension mean subtraction makes the bias a \emph{relative} expert preference rather than a global logit shift.
The bias is broadcast across all tokens in the span, and the prior tensor is frozen so that only the current model parameters train.
$\beta$ controls intervention strength, with $\beta{=}0$ recovering supervised fine-tuning (SFT) with no prior.

\paragraph{Deterministic top-$k$ override (RPB-hard).}
Since $\beta$ never removes the router's own contribution to selection, the extreme of the enforcement axis is defined separately, as a control.
\textbf{RPB-hard} replaces the router logits with the same clipped log-prior as the sole top-$k$ selection score, with gate weights still taken from the unmodified router.
Every token in a Q or A span is then forced through the span's aggregate top-$k$, which removes token-conditioned expert selection.
The override applies at training only, and the router is unmodified at evaluation.

\subsection{Router-L2}
\label{sec:method-l2}

Router-L2 is a router anchoring intervention that uses no sample-conditional prior.
It adds to the standard fine-tuning objective a quadratic penalty that keeps each router weight matrix near its frozen base value:
\begin{equation}
L = L_{\mathrm{SFT}} + \lambda\, \mathrm{mean}_{\ell}\,
\bigl\| W_\ell - W^{\mathrm{base}}_\ell \bigr\|_2^2.
\end{equation}

Only router parameters are anchored: experts, attention, and MLP blocks remain fully trainable, and the router still makes per-token decisions.
Unlike RPB, it uses no sample IDs, span labels, or prior tensors, which makes it the cleanest test of whether the sample-conditional prior accounts for the gain.
The main experiments sweep $\lambda \in \{10^2, 10^3, 10^4\}$ and report $\lambda{=}10^4$ unless noted, with per-setting results in Appendix~\ref{app:lambda-sweep}.

\subsection{Output-Space Router Anchoring}
\label{sec:method-output-anchor}

RPB and Router-L2 leave the router's output distribution unconstrained, the space a direct distillation objective would act on, so we add two span-level output-space anchors against the same frozen base-router prior.
Let $\pi_{x,s,\ell}$ denote the frozen base prior for sample $x$, span $s$, and layer $\ell$, and let $p_{\theta,t,\ell}$ be the current router probability for token $t$ in that span.
The KL-to-base-router baseline adds
\[
\mathcal{L}_{\mathrm{KL\mbox{-}base}}
= \frac{1}{|\mathcal{T}|}\sum_{t \in \mathcal{T}}
D_{\mathrm{KL}}\!\left(\pi_{x(t),s(t),\ell}\,\|\,p_{\theta,t,\ell}\right),
\]
where $\mathcal{T}$ is the set of Q/A tokens with a defined span prior.
The logit-L2-to-base baseline instead compares centered logits:
\begin{align*}
\mathcal{L}_{\mathrm{logit\mbox{-}L2}}
&= \frac{1}{|\mathcal{T}|}\sum_{t \in \mathcal{T}}
\left\|\delta_{t,\ell}\right\|_2^2 , \\
\delta_{t,\ell}
&= \mathrm{center}(z_{t,\ell}) \\
&\quad - \mathrm{center}(\log \pi_{x(t),s(t),\ell}) .
\end{align*}
Here $\mathrm{center}(\cdot)$ subtracts the mean across experts, the same centering applied to the RPB bias, and both objectives are training-time only, matching the bias-free protocol used for RPB.
Unlike RPB, they pull the router output toward the frozen base target through the loss, so the anchoring never enters the top-$k$ selection step.

\begin{table*}[t]
  \centering
  \small
  \begin{tabular}{llll}
    \toprule
    method & role & constrained object & token-conditioned routing \\
    \midrule
    \multicolumn{4}{l}{\emph{Baselines}} \\
    SFT / NoAux     & task loss only                 & none                                 & preserved \\
    LBL             & load-balance baseline          & marginal expert load                 & preserved \\
    \midrule
    \multicolumn{4}{l}{\emph{Base-router-preserving interventions}} \\
    RPB             & soft prior intervention        & router logits toward base prior      & preserved (soft bias) \\
    Router-L2       & parameter-space anchor         & router weights toward base weights   & preserved \\
    KL-to-base      & probability-space anchor       & router distribution toward base      & preserved \\
    Logit-L2        & centered-logit-space anchor    & centered router logits toward base   & preserved \\
    \midrule
    \multicolumn{4}{l}{\emph{Hard-enforcement and shuffled-prior controls}} \\
    RPB-hard        & hard base-prior override       & base prior forces top-$k$            & removed \\
    within-soft     & soft shuffled-prior probe      & shuffled community prior, soft       & preserved (soft bias) \\
    within-hard     & hard within-community override & shuffled prior forces top-$k$        & removed \\
    global-hard     & hard global override           & global shuffled prior forces top-$k$ & removed \\
    \bottomrule
  \end{tabular}
  \caption{Training conditions compared in the main results, grouped by intervention family. \emph{Baselines} fine-tune without (SFT / NoAux) or with marginal-load balancing (LBL); \emph{base-router-preserving interventions} apply different anchoring mechanisms toward the base router; and the \emph{controls} vary the strength of enforcement (RPB-hard) and the content of the prior (the shuffled variants) separately.}
  \label{tab:interventions}
\end{table*}

\section{Experimental Setup}

\paragraph{Models.}
We inspect three base (pretrained) MoE checkpoints: \texttt{moonshotai/Moonlight-16B-A3B}, \texttt{deepseek-ai/DeepSeek-V2-Lite}, and \texttt{Qwen/Qwen3-30B-A3B-Base}.
DeepSeek-V2-Lite and Moonlight-16B-A3B share the same expert geometry: top-6 selection over 64 routed experts, two always-on \emph{shared experts}, and the same per-expert hidden dimension.
They differ in how the router scores those experts, a difference we return to in \S\ref{sec:results-deepseek-main}: Moonlight-16B-A3B uses sigmoid gating with bias-corrected, auxiliary-loss-free top-$k$ selection, whereas DeepSeek-V2-Lite uses softmax scoring with greedy top-$k$.
Qwen3-30B-A3B-Base widens the routed pool to top-8 selection over 128 experts across 48 layers, against top-6 of 64 across 27 layers for the other two, so it varies routing width as well as model family.
It supplies base-router statistics as a frozen probe (Appendix~\ref{sec:appendix-probe}) and is post-trained as a second family in \S\ref{sec:results-qwen}.
Hardware and software configuration are in Appendix~\ref{app:compute}.

\paragraph{Testbed.}
\label{sec:setup-testbed}
Base MoE routers differ in how strongly they separate domains, which bounds the resolution any community-level analysis can achieve.
Moonlight-16B-A3B's base routing is far from uniform, with strong Q/A cross-domain divergence, while DeepSeek-V2-Lite's is diffuse and nearly domain-invariant (Appendix~\ref{app:deepseek-why}), so we center the intervention analysis on the former and treat the latter as a cross-architecture scope condition (Appendix~\ref{app:deepseek-scope}).
Qwen3-30B-A3B-Base enters for the performance claim of \S\ref{sec:results-qwen}, not the community analysis; its base router also separates domains clearly, most sharply on Q spans, though Table~\ref{tab:qwen-base-qa-span} measures this by centroid cosine rather than the $L_1$ statistics used for the other two, so it is not placed on the same sharpness axis.

\paragraph{Datasets.}
\textbf{math20k} and \textbf{coding20k} are 20{,}000-row post-training splits derived from GLM-5.1-Reasoning-1M-Cleaned~\cite{glm51_reasoning_1m_cleaned}: math20k from the \texttt{Math} topical subset and coding20k from the \texttt{main} subset filtered for fenced code blocks.
We use only final completions, so both corpora are (prompt, answer) pairs without an explicit reasoning channel.
Pairing the two domains lets us measure in-domain and out-of-domain performance symmetrically, since each domain's out-of-domain set contains the other.
Source parsing, the coding-block heuristic, and the deterministic quality scorer are in Appendix~\ref{app:corpus}.

\paragraph{Evaluation.}
We evaluate on twelve benchmarks across four task families: math (GSM8K, MATH-500, LiveBench Math), coding (HumanEval, MBPP, LiveBench Coding), general, STEM and professional QA (MMLU, MMLU-STEM, MMLU-Pro, GPQA, GPQA-Diamond), and general reasoning (LiveBench Reasoning).
We report $\mathrm{Avg}_{\mathrm{ID}}$ (in-domain mean) and $\mathrm{Avg}_{\mathrm{OOD}}$ (out-of-domain mean), taking the in-domain set to be the post-training domain and the out-of-domain set its complement.
Aggregate scores use a nine-benchmark subset of the twelve, written $\mathrm{Overall}@9$, which drops three overlapping MMLU and GPQA variants; Appendix~\ref{app:benchmark-suite} lists the suite and Appendix~\ref{app:per-benchmark} reports every benchmark individually.
Routing-state metrics use a held-out slice of the corresponding post-training corpus, disjoint from training tokens.
Unless a table notes otherwise, every condition is trained with three seeds (s42 / s43 / s44) and each checkpoint scored as avg@5 over five evaluation repeats, so tables report seed means with standard deviations (Appendix~\ref{app:seed-protocol}); sampling, decoding, prompt templates, and per-task answer extraction are in Appendix~\ref{app:sampling}.

\paragraph{Interventions.}
\label{sec:setup-interventions}
The conditions of Table~\ref{tab:interventions} share the protocol above and differ only in their auxiliary objective or router-attached bias (Appendix~\ref{app:hyperparameters}).
\textbf{SFT} denotes the task loss alone, with no load-balancing objective, and appears as \textbf{NoAux} in Tables~\ref{tab:anchoring-forms} and~\ref{tab:generality}.

\section{Results}
\label{sec:results}

\subsection{Re-Applied Load Balancing Underperforms Soft Router Anchoring}
\label{sec:results-performance}

None of the evidence here depends on community NMI, whose diagnostic role we take up in \S\ref{sec:results-retention}.

\subsubsection{Main Results: Moonlight-16B-A3B}
\label{sec:results-headline}

On Moonlight-16B-A3B, RPB substantially outperforms both SFT and LBL.
On math20k it reaches 45.77 in-domain accuracy against 31.91 for LBL and 29.44 for SFT, and 19.53 out of domain against 14.97 and 15.65 (Table~\ref{tab:moonlight-summary}).
The gain survives disaggregation: RPB attains the best mean on 21 of the 24 (training-dataset, benchmark) cells (Table~\ref{tab:moonlight-benchmark-full}), and all three exceptions fall on coding20k, where SFT scores higher on MMLU and LiveBench Reasoning and LBL is nominally higher on GPQA-Diamond by 0.81 points, well inside the seed noise.

\begin{table}[t]
  \centering
  \small
  \begin{tabular}{llcc}
    \toprule
    \textbf{Data} & \textbf{Method} & \textbf{ID} & \textbf{OOD} \\
    \midrule
    \multirow{3}{*}{math20k}
      & SFT & 29.44\,{\scriptsize$\pm$0.32} & 15.65\,{\scriptsize$\pm$0.38} \\
      & LBL & 31.91\,{\scriptsize$\pm$0.10} & 14.97\,{\scriptsize$\pm$0.51} \\
      & RPB & \textbf{45.77}\,{\scriptsize$\pm$0.84} & \textbf{19.53}\,{\scriptsize$\pm$0.28} \\
    \midrule
    \multirow{3}{*}{coding20k}
      & SFT & 28.34\,{\scriptsize$\pm$1.36} & 35.83\,{\scriptsize$\pm$0.49} \\
      & LBL & 30.73\,{\scriptsize$\pm$0.13} & 35.56\,{\scriptsize$\pm$0.38} \\
      & RPB & \textbf{41.37}\,{\scriptsize$\pm$0.30} & \textbf{38.70}\,{\scriptsize$\pm$0.11} \\
    \bottomrule
  \end{tabular}
  \caption{Moonlight-16B-A3B in-domain and out-of-domain accuracy over the nine-benchmark suite
  (Appendix~\ref{app:benchmark-suite}; 3 ID / 6 OOD in both blocks), mean $\pm$ standard deviation across three
  training seeds (s42 / s43 / s44) with each seed score an avg@5. RPB improves in-domain adaptation
  without the out-of-domain cost incurred by re-applied LBL.}
  \label{tab:moonlight-summary}
\end{table}

One reading of both effects is that experts outside a token's inherited community are less exposed to the post-training gradient, though we do not measure this directly.

\subsubsection{High Performance Does Not Depend on the Anchoring Form}
\label{sec:results-anchoring}

If the gain depended on the particular form of RPB's bias, anchoring the router in a different space should not reproduce it.
Router-L2 (\S\ref{sec:method-l2}) is the first such test, replacing the sample-conditional logit bias with a loss-integrated penalty on the router weights themselves; on Moonlight-16B-A3B it matches RPB to within $0.6$ points on both corpora, so the improvement is not specific to the bias form.
Both objectives nonetheless leave the router's output distribution unconstrained, so we add the two output-space anchors of \S\ref{sec:method-output-anchor}, which between them cover the natural router-distillation targets, and report all four under an identical protocol (Table~\ref{tab:anchoring-forms}).
They fall within about one point of one another on math20k and half a point on coding20k, against a gap of roughly seven and six points to either baseline, and their ordering does not survive a change of corpus: on coding20k the centered logit-L2 anchor outperforms RPB ($39.81 \pm 0.67$ against $39.59 \pm 0.14$), while on math20k the ordering reverses.
Reaching this level therefore does not require RPB's sample-conditional prior, only that the router be held near its base routing while remaining trainable.

\begin{table}[t]
  \centering
  \small
  \resizebox{\linewidth}{!}{%
  \begin{tabular}{lcc}
    \toprule
    \textbf{Method} & \textbf{math20k} & \textbf{coding20k} \\
    \midrule
    \multicolumn{3}{l}{\emph{Soft router anchoring}} \\
    RPB (span-level logit bias) & \textbf{28.27}\,{\scriptsize$\pm$0.16} & 39.59\,{\scriptsize$\pm$0.14} \\
    Router-L2 (parameter space) & 27.69\,{\scriptsize$\pm$0.28} & 39.35\,{\scriptsize$\pm$0.35} \\
    KL-to-base (probability space) & 27.50\,{\scriptsize$\pm$0.62} & 39.27\,{\scriptsize$\pm$0.33} \\
    Logit-L2 (centered logit space) & 27.24\,{\scriptsize$\pm$0.77} & \textbf{39.81}\,{\scriptsize$\pm$0.67} \\
    \midrule
    \multicolumn{3}{l}{\emph{Baselines}} \\
    NoAux / SFT (no load balancing) & 20.25\,{\scriptsize$\pm$0.29} & 33.34\,{\scriptsize$\pm$0.77} \\
    LBL (re-applied) & 20.61\,{\scriptsize$\pm$0.33} & 33.95\,{\scriptsize$\pm$0.22} \\
    \bottomrule
  \end{tabular}
  }
  \caption{Four anchoring objectives acting in different spaces perform comparably on
  Moonlight-16B-A3B, with Router-L2 at $\lambda{=}10^4$. Values are $\mathrm{Overall}@9$, the mean over the nine-benchmark suite of
  Appendix~\ref{app:benchmark-suite}, reported as mean $\pm$ standard deviation across three training seeds
  (s42 / s43 / s44) with each seed score an avg@5. Best per column in bold. The winner changes
  between the two training corpora, so no single anchoring form is superior on both.}
  \label{tab:anchoring-forms}
\end{table}

\subsubsection{Generality Across Model Family and Corpus}
\label{sec:results-qwen}
\label{sec:results-openr1}

We vary the model family and the corpus in turn, under the same protocol (Table~\ref{tab:generality}).
The first perturbation is a change of MoE family: we post-train Qwen3-30B-A3B-Base on both corpora (upper block).
Re-applied LBL is again the weakest condition on both, so the ordering reproduces on a second family, but only the math20k margin is resolvable at our seed budget: LBL trails RPB by $2.64$ points there, while the $0.16$-point gap on coding20k sits inside the noise.
The advantage over the no-load-balancing baseline does not carry over at all: RPB, Router-L2 and NoAux fall within a band narrower than the seed noise on math20k, and on coding20k NoAux outperforms every other condition, both anchors included.

The second is a change of corpus, since agreement between two splits of one source says little about corpus dependence.
On OpenR1-Math-220k, an independently constructed mathematics corpus yielding roughly 94k examples after the same preprocessing (lower block), re-applied LBL costs $10.20$ points of $\mathrm{Overall}@9$ against RPB, more than twenty times the seed noise, while the gap to the no-load-balancing baseline is $0.96$, small but outside the combined standard deviation.

The two perturbations therefore split the claim: the disadvantage of re-applied LBL survives both; the advantage over unconstrained fine-tuning survives neither.

\begin{table}[t]
  \centering
  \small
  \resizebox{\linewidth}{!}{%
  \begin{tabular}{llcc}
    \toprule
    & \textbf{Method} & \textbf{math20k} & \textbf{coding20k} \\
    \midrule
    \multirow{4}{*}{\rotatebox{90}{\scriptsize Qwen3-30B}}
      & RPB & \textbf{57.58}\,{\scriptsize$\pm$0.74} & 63.13\,{\scriptsize$\pm$0.08} \\
      & Router-L2 & 57.39\,{\scriptsize$\pm$0.40} & 63.93\,{\scriptsize$\pm$0.21} \\
      & NoAux & 56.96\,{\scriptsize$\pm$0.84} & \textbf{64.10}\,{\scriptsize$\pm$0.51} \\
      & LBL & 54.94\,{\scriptsize$\pm$1.87} & 62.97\,{\scriptsize$\pm$0.21} \\
    \midrule
    \multicolumn{2}{l}{\emph{Moonlight-16B-A3B, OpenR1-Math-220k}} & \multicolumn{2}{c}{} \\
      & RPB & \multicolumn{2}{c}{\textbf{41.40}\,{\scriptsize$\pm$0.28}} \\
      & NoAux & \multicolumn{2}{c}{40.44\,{\scriptsize$\pm$0.34}} \\
      & LBL & \multicolumn{2}{c}{31.20\,{\scriptsize$\pm$0.44}} \\
    \bottomrule
  \end{tabular}
  }
  \caption{Generality of the advantage over re-applied LBL. The upper block is Qwen3-30B-A3B-Base
  on math20k and coding20k, and the lower block is Moonlight-16B-A3B on OpenR1-Math-220k. Values are
  $\mathrm{Overall}@9$, mean $\pm$ standard deviation across three training seeds. Re-applied LBL is
  the weakest condition in every block, while the separation between the anchors and the
  no-load-balancing baseline is model- and corpus-dependent.}
  \label{tab:generality}
\end{table}

\subsection{Routing Retention as a Diagnostic of Soft Router Anchoring}
\label{sec:results-retention}

We now ask which level of routing retention tracks the performance ordering.

\subsubsection{Community Structure as a Footprint of Soft Enforcement}
\label{sec:results-hierarchy}
\label{sec:results-footprint}

The three levels of \S\ref{sec:method-graph} do not separate the methods equally (Table~\ref{tab:retention-routerl2}).
SFT, LBL and RPB all fall in the same TVD and JSD range, so per-expert utilization carries no information about which intervention was applied.
Top-$k$ overlap does separate LBL from the rest, but it places SFT between the two anchors on coding20k (0.482, against 0.481 for RPB and 0.486 for Router-L2), so only community-level retention isolates the soft-anchoring family on both corpora.

Separating the methods does not, however, make community NMI the quantity to maximize.
It saturates early under a $\beta$ sweep, and replacing the soft bias with the hard assignment of \S\ref{sec:method-rpb} pushes it above the soft maximum while performance falls well below it, on coding20k even below unanchored fine-tuning (Appendix~\ref{sec:appendix-beta-operating}), so it is a footprint left by soft enforcement at training time rather than an optimization target.

\subsubsection{Interventional Controls: Soft versus Hard Enforcement}
\label{sec:results-shuffled}

\begin{figure*}[t]
\centering
\includegraphics[width=\textwidth]{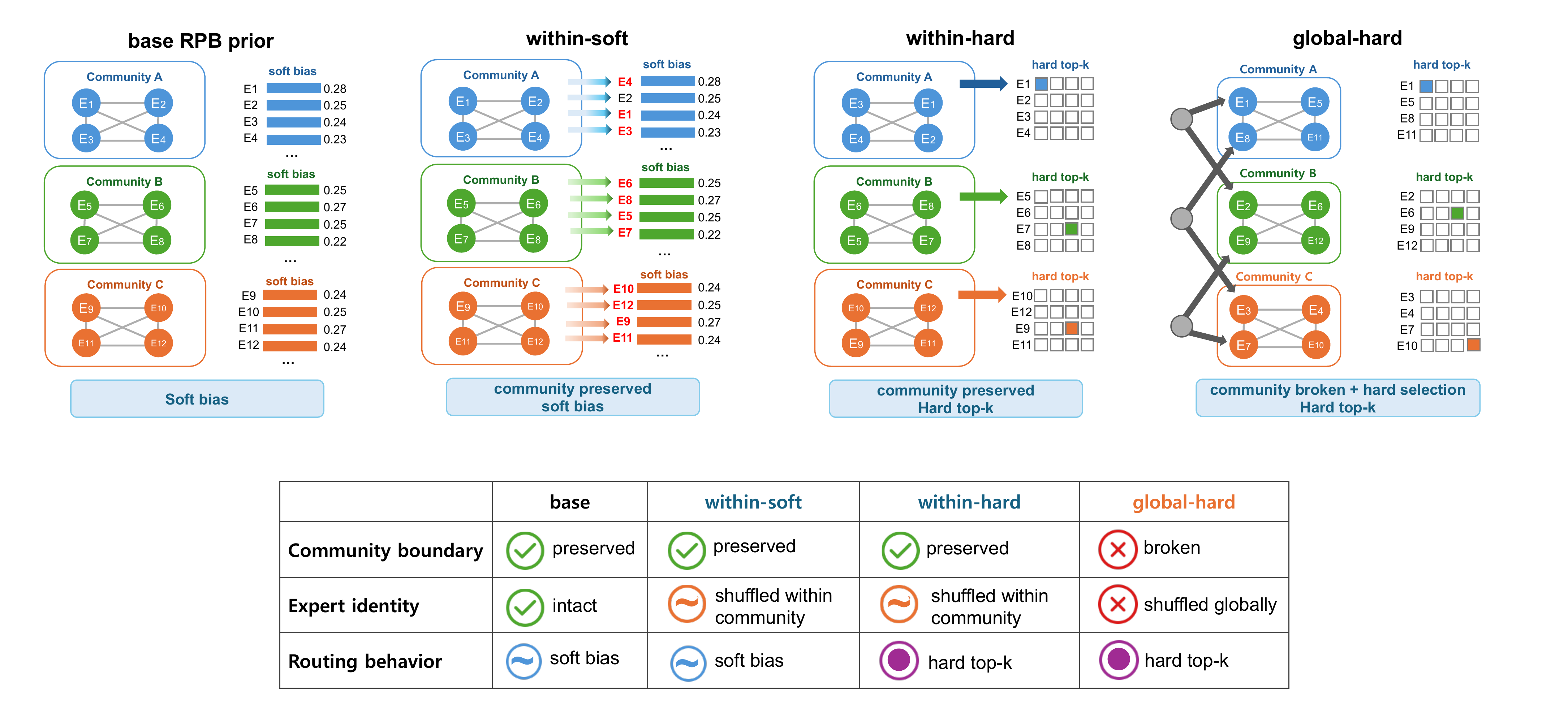}
\caption{
Three shuffle controls applied to the base RPB prior over experts $E_i$.
In \textbf{within-soft}, expert identities are permuted only within each detected community, while the shuffled prior is still applied as a soft router bias, so community structure is preserved and token-conditioned routing remains soft.
In \textbf{within-hard}, the same within-community permutation is converted into hard top-$k$ expert selection, preserving community membership but removing soft router choice.
In \textbf{global-hard}, expert identities are permuted across communities and then used for hard top-$k$ selection, so the community structure is broken in addition to soft routing being replaced by hard assignment.
}
\label{rpb-shuffle}
\end{figure*}

A positive RPB result does not on its own separate the content of the prior from the strength with which it is enforced.
We vary the two axes independently (Figure~\ref{rpb-shuffle}) with three controls: \textbf{within-soft} (within-community shuffle, soft logit bias), \textbf{within-hard} (within-community shuffle, hard top-$k$ override), and \textbf{global-hard} (across-community shuffle, hard override).
The within-community shuffle is a substantial perturbation rather than a nominal one, changing 46--60\% of the prior's top-$k$ entries by an amount comparable in magnitude to the router logits themselves (Appendix~\ref{sec:appendix-shuffle-validation}).

\begin{table}[t]
  \centering
  \small
  \resizebox{\linewidth}{!}{%
  \begin{tabular}{lcccc}
    \toprule
    & \multicolumn{2}{c}{\textbf{math20k}} & \multicolumn{2}{c}{\textbf{coding20k}} \\
    \cmidrule(lr){2-3}\cmidrule(lr){4-5}
    \textbf{Condition} & \textbf{Overall@9} & \textbf{NMI} & \textbf{Overall@9} & \textbf{NMI} \\
    \midrule
    RPB (unshuffled reference) & \textbf{28.27} & 0.631 & \textbf{39.59} & 0.629 \\
    within-soft & 27.23 & 0.632 & 38.96 & 0.632 \\
    within-hard & 18.77 & 0.634 & 24.87 & 0.636 \\
    global-hard & 3.59 & \textbf{0.642} & 17.96 & \textbf{0.650} \\
    \bottomrule
  \end{tabular}
  }
  \caption{Shuffled-prior controls on Moonlight-16B-A3B. Holding community labels fixed and switching
  enforcement from soft to hard (within-soft $\to$ within-hard) costs more than scrambling the labels
  while keeping enforcement soft (RPB $\to$ within-soft). Community NMI moves in the opposite direction
  to performance, since the global-hard control attains the highest retention on both corpora while
  performance falls sharply. $\mathrm{Overall}@9$ columns are three-seed means (s42 / s43 / s44); NMI
  columns are three probed seeds for RPB and two (s42 / s43) for the shuffled controls, following
  Appendix~\ref{app:seed-protocol}. Per-benchmark breakdown in Appendix~\ref{app:per-benchmark-shuffle}.}
  \label{tab:shuffle-summary}
\end{table}

Three observations follow from Table~\ref{tab:shuffle-summary}.
First, \emph{within-soft stays within about one point of RPB}, with community NMI matched as well (Table~\ref{tab:shuffle-summary}), so the prior need not name individual base experts and community-level anchoring suffices.
Second, \emph{within-hard reduces performance sharply despite the same community-preserving prior content}, leaving enforcement strength as the only axis that changed and hence as the operative one.
Third, \emph{global-hard reduces performance further still}, once community membership is corrupted as well; its community NMI is nonetheless the highest of the four conditions, which reflects the limited resolution of NMI on weakly clustered graphs rather than genuine retention, and is why we validate the community object against a marginal-preserving null model and report modularity $Q$ alongside it (Appendix~\ref{app:null-model}).

\subsubsection{Scope Condition: DeepSeek-V2-Lite}
\label{sec:results-deepseek-main}

The diagnostic presupposes a base router with community structure to retain, and DeepSeek-V2-Lite is a model where that fails: its routing is close to uniform and nearly domain-invariant (\S\ref{sec:setup-testbed}); we report this as a measured property of the base checkpoint rather than attribute it to any single design choice.

Community NMI then has little structure to compare across conditions and fails to separate the interventions as it does on Moonlight-16B-A3B (Appendix~\ref{app:deepseek-scope}).
Performance still improves under RPB there, and the soft-against-hard dissociation still reproduces, so we read the flat NMI as the diagnostic behaving as specified rather than as a failure of the intervention, and state it only for base models whose routing is non-uniform enough to induce community structure.

\section{Discussion}

The four anchoring objectives act on router parameters, on router logits, or on the routing distribution, yet perform comparably, and what they share is that each keeps the router close to its base routing while leaving it trainable.
LBL, constraining only the marginal load, does not reach that level, while the hard variants push the constraint past it, reaching comparable or higher community NMI while performance declines.
That the space of the anchor makes no consistent difference while its softness makes a large one places the effect in the strength of the constraint rather than in the quantity constrained.
The ST-MoE z-loss reaches the same conclusion from the opposite direction, regularizing the router with no reference to the base model and trailing even unanchored fine-tuning on both corpora (Appendix~\ref{app:zloss}).

The dissociation between community NMI and performance carries a warning beyond MoE routing.
Inside the soft family, retained community structure covaries with accuracy closely enough to suggest an objective worth maximizing, yet hard enforcement attains retention at or above the soft maximum while performance drops sharply.
A proxy of the form ``distance from the base model'' is therefore informative only over the range in which it was calibrated, and optimizing it directly is what pushes it outside that range.

At fixed hard enforcement, shuffling expert identities across communities rather than within one costs a further 15.2 points on math20k and 6.9 on coding20k, so post-training reuses specialization at the granularity at which experts collaborate.

Two design constraints follow.
The anchoring target should be the inherited co-activation structure rather than the per-expert load distribution, since flattening the latter removes the property that supports out-of-domain retention; and the anchoring should stay soft, since the interventional controls locate the failure at enforcement strength rather than at prior content.
Methods that freeze or hard-select experts during adaptation~\cite{wang2024let,Li2025DynamicES,eo2025mixture} fix their enforcement strength by construction, so graded variants are worth testing against the range Router-L2 spans.
What determines whether the diagnostic applies is how sharply the base router already separates domains rather than the expert architecture: shared-expert designs~\cite{dai-etal-2024-deepseekmoe,guo2025advancing} appear on both sides of our comparison, so that design choice does not by itself settle the question.

\section{Related Work}

\paragraph{MoE fine-tuning and routing control.}
Sparse MoE pretraining stabilizes the expert pool through utilization-oriented routing objectives~\cite{Lepikhin2020GShardSG,Fedus2021SwitchTS,Zhou2022MixtureofExpertsWE}, most commonly an auxiliary load-balancing loss (LBL), with refinements such as global-batch variants~\cite{qiu2025demons} and auxiliary-loss-free or dynamic-bias balancing~\cite{Wang2024AuxiliaryLossFreeLB,DeepSeekAI2024DeepSeekV3TR}; all of these target pretraining dynamics, where no inherited routing exists yet to preserve.
The closest work to ours instead asks which experts to update, freeze, or select during adaptation: ESFT~\cite{wang2024let} freezes experts by relevance, DES-MoE~\cite{Li2025DynamicES} targets dynamic specialization, and MoCE~\cite{eo2025mixture} routes within a selected group, close in spirit to soft anchoring.
Practitioner toolchains have converged on similar empirical choices, dropping load balancing or router training altogether during post-training~\cite{nemo-megatron-bridge,unsloth-moe}.
We complement both lines by targeting the router itself, constraining how routing changes during post-training rather than reshaping the expert pool.

\paragraph{Expert specialization.}
A parallel line treats experts as \emph{specialized}.
Architectural designs separate shared and fine-grained routed experts~\cite{guo2025advancing,dai-etal-2024-deepseekmoe}; Moonlight-16B-A3B and DeepSeek-V2-Lite both adopt that design, so the contrast in \S\ref{sec:results-deepseek-main} turns on their router scoring rules rather than on this axis.
Probe studies find semantic and multilingual signal in routing~\cite{Lo2024ACL,bai-etal-2025-understanding}, and frozen-checkpoint analyses target compression~\cite{Lu2024NotAE,Hu2026MosaicPA}.
A complementary view treats expert \emph{collaboration}, meaning which experts co-activate on the same token, as the analysis target, through co-activation matrices and graph-structured MoE~\cite{Tang2026SpecializationTC,NguyenNhat2025ModelingEI}.
We adopt this collaboration view but treat the co-activation graph as a \emph{state variable} whose retention is measurable after post-training, using standard network-science tools: modularity $Q$~\cite{Newman2006ModularityAC} with Louvain~\cite{Blondel2008FastUO} and Leiden~\cite{Traag2018FromLT} community detection.

\section{Conclusion}

Re-applying the pretraining load-balancing loss degrades downstream performance relative to leaving the inherited routing softly anchored, and the effect survives a change of model family and of training corpus.
Four anchoring objectives acting in different spaces perform comparably on Moonlight-16B-A3B, so \textbf{RPB is one competitive implementation of soft router anchoring rather than a uniquely necessary one}.
Subject to two boundaries, that the advantage over a no-load-balancing baseline is model-dependent and that community NMI is informative only where the base router carries community structure to begin with, these findings move the target of MoE post-training stabilization from re-imposing uniform expert utilization to preserving the routing the base model has already learned.

\section*{Limitations}
The community-mediated evidence is Moonlight-centered.
We state the diagnostic with a scope condition rather than as a general property because of DeepSeek-V2-Lite's diffuse routing distribution, and on Qwen3-30B-A3B-Base the diagnostic and the performance claim come apart, since the ordering against re-applied LBL is preserved while the community-level reading is not the operative evidence there.
The within-community shuffle result should likewise be read as a statement about the communities Louvain detects in Moonlight-16B-A3B, not as evidence that they are dense, semantically coherent modules, since we do not characterize their internal composition.
Establishing how widely the community reading applies would require base models sampled across a spectrum of routing sharpness, and further MoE families to test whether the same footprint appears under different community structures; neither is within our scope.

The comparison against a no-load-balancing baseline is the weaker half of our performance result and should not be read as stronger than it is.
Soft anchoring outperforms it by a wide margin on Moonlight-16B-A3B and by a small but reliable one on OpenR1-Math-220k, yet on Qwen3-30B-A3B-Base the two are indistinguishable on math20k and the no-load-balancing baseline is superior on coding20k.
We report the reversal rather than restricting the comparison, but we cannot presently say which property of a base model predicts the direction of the effect.

The fine-tuning data are intentionally controlled 20k-example splits from a single source corpus, GLM-5.1-Reasoning-1M-Cleaned.
This design makes the in-domain and out-of-domain contrast between math and code symmetric and keeps the routing interventions comparable, but it does not establish that the same hyperparameter range or effect size will hold under larger or more heterogeneous instruction mixtures.

Router-L2 shows that the improvement is not specific to an additive logit bias, and the output-space anchors cover the direct-distillation alternative.
None of the four objectives separates anchoring strength from anchoring target within a single objective.
Each fixes a target and varies strength, so we cannot rule out that a different target would move the point at which adaptation and retention are best traded off.
We also do not have a criterion that predicts which anchoring form will lead on a given corpus, only the observation that the ordering changes across corpora.

We focus on non-reasoning base models, with math20k and coding20k as correspondingly non-reasoning QA splits.
Reasoning-trace post-training with long chains of thought is not covered, and the structural retention dynamics may differ when the gradient signal is heavily weighted toward long-form generation.
In such settings the relevant object may not be a static co-activation graph but \emph{sequential routing transition communities}, defined over the time-ordered pattern of expert-set transitions across reasoning steps.
Constructing those transition graphs and re-running the shuffled-control protocol against them is a natural next step.

\bibliography{custom}

\twocolumn[%
  \noindent{\Large\textbf{Appendices}}\par
  \vspace{1.0em}
]
\appendix

\begin{table*}[t]
  \centering
  \small
  \setlength{\tabcolsep}{19pt}
  \begin{tabular}{lcccc}
    \toprule
    & \multicolumn{2}{c}{\textbf{math20k}} & \multicolumn{2}{c}{\textbf{coding20k}} \\
    \cmidrule(lr){2-3}\cmidrule(lr){4-5}
    \textbf{$\beta$ / variant} & \textbf{Community NMI} & \textbf{Overall}@9 & \textbf{Community NMI} & \textbf{Overall}@9 \\
    \midrule
    0 (SFT) & 0.575 & 20.25 & 0.588 & 32.68 \\
    0.5 constant & 0.624 & 27.41 & 0.630 & 38.37 \\
    0.75 constant & 0.629 & 28.11 & 0.629 & \textbf{40.04} \\
    1.0 constant & 0.631 & \textbf{28.44} & 0.629 & 39.66 \\
    RPB-hard & 0.634 & 24.08 & 0.642 & 31.54 \\
    \bottomrule
  \end{tabular}
  \caption{Operating range of RPB across $\beta$ on Moonlight-16B-A3B, with the hard end of the sweep (RPB-hard) for contrast. Community NMI is the layer-mean community-level normalized mutual information against the base partition, and $\mathrm{Overall}@9$ is the nine-benchmark aggregate of Appendix~\ref{app:benchmark-suite}. Single training seed (s42), each score an avg@5 over five evaluation repeats; all comparisons drawn from this table are therefore seed-matched. The $\beta{=}0$ row is the single-seed counterpart of the NoAux baseline of Table~\ref{tab:anchoring-forms}, whose three-seed mean differs slightly, so the two should not be read against each other.}
  \label{tab:beta-operating-range}
\end{table*}

\section{Further Controls: $\beta$ Range and z-loss}
\subsection{$\beta$ Operating Range and the Hard-Assignment Ceiling}
\label{sec:appendix-beta-operating}

Table~\ref{tab:beta-operating-range} sweeps $\beta$, which controls the strength of the RPB prior.
As $\beta$ rises, community NMI rises and saturates by $\beta{=}0.5$, while performance follows a unimodal shape peaking near $\beta{=}0.75$ to $1.0$.
The hard end of the sweep (\textbf{RPB-hard}) uses the same prior as the sole top-$k$ selection score, with contribution weights still computed from router logits.
It raises community NMI slightly above the soft maximum yet reduces performance on both corpora, falling below even $\beta{=}0$ on coding20k, which is the dissociation \S\ref{sec:results-footprint} builds on.

\subsection{Router-L2 $\lambda$ Sensitivity}
\label{app:lambda-sweep}

Table~\ref{tab:router-l2-lambda} completes the sweep announced in \S\ref{sec:method-l2}.
Across two orders of magnitude the anchoring strength moves the operating point only slightly: the three settings span 0.25 points on math20k, inside the seed spread of the $\lambda{=}10^4$ row, and 1.0 point on coding20k.
The ordering also reverses between corpora, with the weakest anchor best on coding20k and the strongest best on math20k, which is the same corpus-dependence the four anchoring objectives show in \S\ref{sec:results-anchoring}.
We report $\lambda{=}10^4$ in the main tables because it is the only setting trained with three seeds, not because it is the sweep optimum.

\begin{table}[t]
  \centering
  \small
  \begin{tabular}{lcc}
    \toprule
    \textbf{$\lambda$} & \textbf{math20k} & \textbf{coding20k} \\
    \midrule
    $10^2$ & 27.44$^{\dagger}$ & \textbf{40.35}$^{\dagger}$ \\
    $10^3$ & 27.51$^{\dagger}$ & 39.64$^{\dagger}$ \\
    $10^4$ & \textbf{27.69}\,{\scriptsize$\pm$0.28} & 39.35\,{\scriptsize$\pm$0.35} \\
    \bottomrule
  \end{tabular}
  \caption{Router-L2 anchoring strength on Moonlight-16B-A3B, reported as $\mathrm{Overall}@9$. The $\lambda{=}10^4$ rows are three-seed means $\pm$ sample standard deviation (s42 / s43 / s44); $^{\dagger}$the two weaker settings are single-seed (s42), so the bolded coding20k entry should not be read as a significant win. Each seed score is itself an avg@5 over five evaluation repeats.}
  \label{tab:router-l2-lambda}
\end{table}

\subsection{ST-MoE z-loss}
\label{app:zloss}

Table~\ref{tab:stmoe-zloss} reports the ST-MoE z-loss~\cite{zoph2022st}, included as a baseline regularizer drawn from the MoE literature rather than as an anchoring objective.
It penalizes large router logits to stabilize training and carries no reference to the base router.
Its rows come from a separate set of runs, so we read the direction of the gaps below rather than their size.

\begin{table}[t]
  \centering
  \small
  \begin{tabular}{ll}
    \toprule
    \textbf{Setting} & \textbf{ST-MoE Avg} \\
    \midrule
    Moonlight math20k & 16.33 \\
    DeepSeek math20k & 9.59 \\
    Moonlight coding20k & 30.05 \\
    DeepSeek coding20k & 17.57 \\
    \bottomrule
  \end{tabular}
  \caption{ST-MoE z-loss~\cite{zoph2022st} results on Moonlight-16B-A3B and DeepSeek-V2-Lite, included as a baseline regularizer drawn from the MoE literature rather than as an anchoring objective. Values are the twelve-benchmark mean at a single training seed (s42), each score an avg@5 over five evaluation repeats, so they are on the same scale as the \textbf{Avg} column of Table~\ref{tab:moonlight-benchmark-full} rather than the $\mathrm{Overall}@9$ of the main tables.}
  \label{tab:stmoe-zloss}
\end{table}

In every case the z-loss falls below RPB, and on Moonlight-16B-A3B it also falls below unanchored fine-tuning on both corpora, whether the SFT figure is taken over the nine-benchmark suite or over all twelve.
The negative result supports the broader point, since it shows that regularizing the router is not sufficient on its own: the regularizer has to act toward the inherited routing structure rather than toward a generic property of the routing distribution.

\section{Validity and Robustness of the Community Structure}

\subsection{Marginal-Preserving Null Model}
\label{app:null-model}

Table~\ref{tab:null-model} verifies that the detected communities are not merely artifacts of marginal expert usage.
We compare each observed co-activation graph against a marginal-preserving null model, namely 100 multinomial-resampled graphs per \texttt{(run, cluster, layer)} whose expected edge weights are proportional to the product of expert marginals, which preserves each expert's marginal mass while removing pairwise co-selection.
Real graphs beat this null on modularity in 100\% of cells for every condition, with $Q$ margins of 0.132 to 0.210.
The NMI margins are smaller throughout, at 0.037 to 0.091, which reinforces that community NMI is diagnostic rather than explanatory.
We use the null model only as a validity check for the community object, not as a method-ranking metric.

\begin{table}[t]
  \centering
  \small
  \begin{tabular}{llcc}
    \toprule
    \textbf{Dataset} & \textbf{Method} & \textbf{$\Delta Q$} & \textbf{$\Delta$NMI} \\
    \midrule
    \multirow{4}{*}{math20k}
      & SFT      & 0.140 & 0.053 \\
      & LBL      & 0.210 & 0.048 \\
      & RPB      & 0.132 & 0.050 \\
      & RPB-hard & 0.140 & 0.039 \\
    \midrule
    \multirow{4}{*}{coding20k}
      & SFT      & 0.149 & 0.052 \\
      & LBL      & 0.174 & 0.091 \\
      & RPB      & 0.132 & 0.046 \\
      & RPB-hard & 0.161 & 0.037 \\
    \bottomrule
  \end{tabular}
  \caption{Marginal-preserving null-model comparison on Moonlight-16B-A3B. $\Delta Q$ and $\Delta$NMI are the margins of the observed co-activation graph over 100 multinomial-resampled null graphs per \texttt{(run, cluster, layer)}, at edge fraction $\rho=0.10$ and a single probed seed (s42). Every condition beats the null on modularity in 100\% of cells, and the modularity margin is consistently larger than the NMI margin, which is why we read community NMI as a diagnostic rather than a sufficient explanation. The margins do not order the methods the way performance does, so this check validates the community object rather than ranking the interventions.}
  \label{tab:null-model}
\end{table}

\subsection{Threshold and Detector Sweeps}
\label{app:sweeps}

Table~\ref{tab:sweeps} varies the two free choices in the community construction of \S\ref{sec:method-graph}.
The edge fraction $\rho$ moves the absolute level of community NMI substantially, since retaining fewer edges leaves a sparser graph whose partition is easier to reproduce, but it leaves the method ordering RPB $\approx$ Router-L2 $>$ SFT $>$ LBL unchanged at every setting on both corpora.
Substituting Leiden for Louvain at otherwise-matched parameters changes no value by more than $0.009$, and dispersion across detector seeds $\{0,1,2,42\}$ averages $0.031$ community NMI and never exceeds $0.050$, so it stays below the RPB-to-LBL gap at every $\rho$ in the sweep.
The separate sweep behind the within-community shuffle, over $\rho \in \{0.05, 0.10, 0.15, 0.20\}$ and permutation seeds $\{0,1,2\}$, keeps within-community preservation at $0.986$ community NMI on average, never below $0.957$ and with standard deviation $0.011$ across all combinations, so the perturbation of \S\ref{sec:results-shuffled} preserves community membership by construction rather than by chance.

\begin{table}[t]
  \centering
  \small
  \begin{tabular}{llccc}
    \toprule
    & & \multicolumn{3}{c}{\textbf{Community NMI at $\rho$}} \\
    \cmidrule(lr){3-5}
    \textbf{Dataset} & \textbf{Method} & \textbf{0.05} & \textbf{0.10} & \textbf{0.20} \\
    \midrule
    \multirow{4}{*}{math20k}
      & SFT       & 0.729 & 0.580 & 0.410 \\
      & LBL       & 0.709 & 0.547 & 0.393 \\
      & RPB       & \textbf{0.752} & \textbf{0.634} & \textbf{0.493} \\
      & Router-L2 & 0.750 & 0.628 & 0.488 \\
    \midrule
    \multirow{4}{*}{coding20k}
      & SFT       & 0.734 & 0.590 & 0.444 \\
      & LBL       & 0.689 & 0.541 & 0.408 \\
      & RPB       & 0.748 & 0.629 & \textbf{0.497} \\
      & Router-L2 & \textbf{0.752} & \textbf{0.632} & 0.493 \\
    \bottomrule
  \end{tabular}
  \caption{Sensitivity of community NMI to the edge-fraction threshold on Moonlight-16B-A3B, at a single probed seed (s42) with Louvain at resolution $\gamma=1.0$. The absolute level falls as $\rho$ grows and the graph densifies, but the ordering of the four methods is preserved at every threshold. Leiden at matched parameters agrees with Louvain to within $0.009$ on every cell, and dispersion over detector seeds $\{0,1,2,42\}$ averages $0.031$ and never exceeds $0.050$. The main text uses $\rho = 0.10$. Because these are single-seed values they differ slightly from the three-seed means of Table~\ref{tab:retention-routerl2}. Bold marks the best mean per column within each dataset block.}
  \label{tab:sweeps}
\end{table}

\begin{table*}[t]
  \centering
  \small
  \resizebox{\textwidth}{!}{%
  \begin{tabular}{llcccc}
    \toprule
    \textbf{Model} & \textbf{Span} & \textbf{math--code cos.} & \textbf{math--general cos.} & \textbf{code--general cos.} & \textbf{eff.\ expert frac.} \\
    \midrule
    Moonlight-16B-A3B & Q & 0.438 & 0.790 & 0.362 & 0.202--0.260 \\
    Moonlight-16B-A3B & A & 0.788 & 0.943 & 0.745 & 0.245--0.281 \\
    Qwen3-30B-A3B-Base & Q & 0.409 & 0.697 & 0.378 & 0.228--0.257 \\
    Qwen3-30B-A3B-Base & A & 0.561 & 0.800 & 0.525 & 0.348--0.404 \\
    \bottomrule
  \end{tabular}
  }
  \caption{Q/A span routing structure in Moonlight-16B-A3B and Qwen3-30B-A3B-Base, measured from mean top-$k$ activation frequency (\texttt{P\_topk}) on the math/code/general probe clusters. Cosine similarity is computed between domain centroids after averaging over samples and layers, so lower values indicate stronger domain separation. Effective expert fraction is $\exp(H)/E$ averaged over the listed domains, where $H$ is the entropy of the mean top-$k$ activation frequency over the $E$ routed experts, so larger values indicate broader expert usage. All quantities are measured from the frozen base checkpoints, so no training seeds are involved.}
  \label{tab:qwen-base-qa-span}
\end{table*}

\section{Base-Router Heterogeneity Probes}
\label{sec:appendix-probe}

To check whether the Q/A split used by RPB is specific to Moonlight-16B-A3B and DeepSeek-V2-Lite or also appears in a newer MoE family, we additionally probed Qwen3-30B-A3B-Base.
In this appendix the model serves only as a frozen base router, downloaded directly from HuggingFace, and its post-training results are reported separately in \S\ref{sec:results-qwen}.
It is a non-reasoning probe in our setup, since no probe row carries a separate reasoning trace, so every sequence decomposes into the Q and A spans of \S\ref{sec:method-rpb} alone.
Table~\ref{tab:qwen-base-qa-span} reports the two spans for Moonlight-16B-A3B and Qwen3-30B-A3B-Base.
Q tokens separate the math, code, and general probe clusters more sharply than A tokens do in both models, on all three pairwise cosine similarities, and code is the most separated cluster on Moonlight-16B-A3B.
The pattern holds on Qwen3-30B-A3B-Base despite its larger routed pool of 128 experts, although its A spans retain more domain separation and draw on a broader expert fraction than Moonlight-16B-A3B's, so the split RPB conditions on is not an artifact of a single model family.

\section{DeepSeek-V2-Lite as a Scope Condition for the Diagnostic}
\label{app:deepseek-scope}
\subsection{Cross-Architecture Robustness Check}

DeepSeek-V2-Lite has a considerably more diffuse routing distribution than Moonlight-16B-A3B, which makes it a useful cross-architecture robustness test: the performance claim and the community-level diagnostic can be checked separately on it.
RPB still improves performance, most clearly out of domain on math20k, but community NMI does not separate the methods at all, with all four fine-tuning conditions clustering near NMI $\approx 0.49$ (Table~\ref{tab:deepseek-robustness}).
Appendix~\ref{app:deepseek-why} analyzes the underlying mechanism, namely a diffuse and nearly domain-invariant base routing distribution.

\begin{table}[t]
  \centering
  \footnotesize
  \setlength{\tabcolsep}{3.5pt}
  \resizebox{\linewidth}{!}{%
  \begin{tabular}{llllll}
    \toprule
    \textbf{Dataset} & \textbf{Method} & \textbf{ID} & \textbf{OOD} & \textbf{Avg} & \textbf{Comm. NMI} \\
    \midrule
    \multirow{5}{*}{math20k} & Base & 3.77 & 17.52 & 14.08 & 1.000 \\
      & SFT & 13.55 & 14.42 & 14.21 & 0.490 \\
      & LBL & 13.43 & 15.69 & 15.12 & 0.487 \\
      & RPB & 14.11 & 16.79 & 16.12 & 0.498 \\
      & RPB-hard & 11.47 & 8.76 & 9.44 & 0.503 \\
    \multirow{5}{*}{coding20k} & Base & 17.01 & 13.11 & 14.08 & 1.000 \\
      & SFT & 21.92 & 21.31 & 21.46 & 0.488 \\
      & LBL & 21.86 & 20.99 & 21.21 & 0.493 \\
      & RPB & 22.88 & 22.49 & 22.58 & 0.497 \\
      & RPB-hard & 9.49 & 19.40 & 16.92 & 0.495 \\
    \bottomrule
  \end{tabular}
  }
  \caption{DeepSeek-V2-Lite cross-architecture robustness check. Single training seed (s42), each score an avg@5 over five evaluation repeats. Unlike the Moonlight-16B-A3B tables, the \textbf{ID}, \textbf{OOD}, and \textbf{Avg} columns here aggregate all twelve benchmarks rather than the nine-benchmark suite, so they cannot be recomputed from Table~\ref{tab:deepseek-benchmark-detail}, which lists the nine. \textbf{Comm. NMI} is community NMI against the base partition.}
  \label{tab:deepseek-robustness}
\end{table}

The dependence on soft, token-conditioned routing remains visible even without community-level separation.
Under hard span-level control (RPB-hard), performance drops sharply, from $16.12$ to $9.44$ overall on math20k and from $22.58$ to $16.92$ overall on coding20k, with in-domain accuracy falling from $22.88$ to $9.49$.

Two readings follow.
The community-mediated channel is model-dependent, since Moonlight-16B-A3B supports the full community dissociation while DeepSeek-V2-Lite shows that performance gains can occur without strong community NMI separation when the base routing distribution is diffuse.
The dependence on soft rather than hard enforcement, by contrast, holds on both post-trained models, though two models establish a consistent pattern rather than a general property.

\subsection{Why Community NMI Lacks Separation on DeepSeek-V2-Lite}
\label{app:deepseek-why}

The absence of NMI separation on DeepSeek-V2-Lite (Appendix~\ref{app:deepseek-scope}, Table~\ref{tab:deepseek-robustness}) reflects three properties of its base routing distribution.
\emph{(i) Diffuse routing distribution.} The base routing sharpness $\|p - U\|_1$, the $L_1$ distance between the layer-averaged expert distribution $p$ and the uniform distribution $U$ over routed experts, is 0.1399 on math20k for DeepSeek-V2-Lite against 0.5445 for Moonlight-16B-A3B, so its base distribution sits much closer to uniform.
\emph{(ii) Low cross-domain divergence.} $\|p_{\rm math} - p_{\rm code}\|_1 = 0.0724$ for DeepSeek-V2-Lite against 0.3735 for Moonlight-16B-A3B, so its routing preferences are nearly domain-invariant.
\emph{(iii) Router scoring rule.} The two base routers share their expert geometry, both using top-6 of 64 routed experts alongside two shared experts at the same per-expert hidden dimension, and differ instead in how they score those experts: Moonlight-16B-A3B uses sigmoid gating with bias-corrected, aux-loss-free top-$k$ selection, while DeepSeek-V2-Lite uses softmax scoring with greedy top-$k$.
We did not isolate this difference experimentally, so we report it as a candidate explanation for (i) and (ii) rather than a demonstrated cause.

The graph-level consequence is that the DeepSeek-V2-Lite co-activation graphs are naturally more uniform and less clustered, and Louvain on a near-uniform graph cannot return a partition sharply different from the base one whatever the fine-tuning method.
All four fine-tuned DeepSeek-V2-Lite variants land within 0.02 of one another at NMI $\approx$ 0.49, while the Moonlight-16B-A3B variants span a much wider band, so community NMI cannot separate methods when the base community structure is itself weak.

\section{Prior-Side Validation of Shuffled-Prior Controls}
\label{sec:appendix-shuffle-validation}

\paragraph{2$\times$2 design.}
The three controls used in \S\ref{sec:results-shuffled} sit on a 2$\times$2 grid of \emph{shuffle scope} against \emph{enforcement strength}.
\textbf{Within-community shuffle} permutes expert identities inside each detected base community, so a token that would have been routed to community $C$ is still routed inside $C$ and only its identity within $C$ is scrambled.
\textbf{Global shuffle} permutes expert identities across all communities, so both community membership and within-community identity are corrupted at once.
Each scope is paired with a \textbf{soft} application, where the shuffled prior enters as a soft logit bias and token-conditioned routing stays intact, or a \textbf{hard} application, where the shuffled prior forces the Q/A-span aggregate top-$k$ and removes token-conditioned routing.
The three reported cells, \emph{within-soft}, \emph{within-hard}, and \emph{global-hard}, together isolate the soft against hard switch at fixed community preservation (within-soft against within-hard) and the within against global scope at fixed hardness (within-hard against global-hard).

\paragraph{Outcome statistic vs.\ intervention strength.}
Two measurements must not be conflated here.
The bias-free probe of the eventual checkpoint shows within-soft and unshuffled RPB landing at near-identical top-$k$ overlap to base, which is an \emph{outcome statistic} about the trained model's routing behavior.
The prior-side measurements below instead report \emph{intervention strength}, the perturbation applied to the prior tensor before training ever begins.

\begin{table*}[t]
  \centering
  \small
  \resizebox{\textwidth}{!}{%
  \begin{tabular}{lllllll}
    \toprule
    \textbf{Dataset} & \textbf{Variant} & \textbf{prior top-$k$ overlap} & \textbf{top-$k$ change rate} & \textbf{effective log-bias $|\Delta b|$ p90} & \textbf{clamp upper-bound frac.} & \textbf{Verdict} \\
    \midrule
    math20k & within-community & 0.539 & 0.460 & 0.471 & 0.021 & meaningful intervention \\
    math20k & global & 0.104 & 0.896 & 1.455 & 0.021 & strong negative control \\
    coding20k & within-community & 0.397 & 0.604 & 0.822 & 0.011 & meaningful intervention \\
    coding20k & global & 0.094 & 0.906 & 1.581 & 0.011 & strong negative control \\
    \bottomrule
  \end{tabular}
  }
  \caption{Prior-side intervention strength of the within-community and global shuffles, measured on the RPB prior tensor before any training, so the values do not depend on training seeds. The overlap and change-rate columns are complementary by construction. The remaining columns are the 90th percentile of the per-element change in the effective log-prior bias and the fraction of entries held at the clamp upper bound.}
  \label{tab:shuffle-strength}
\end{table*}

\subsection{Prior-Side Intervention Strength}
\label{app:shuffle-strength}

The validation script compares the \emph{original} unshuffled RPB prior against the \emph{shuffled} prior at a fixed clamp and $\beta$ configuration.
The prior tensor is \texttt{prior\_qa}, of shape $N_{\mathrm{samples}} \times 2 \times L \times E$ over the Q and A spans (Appendix~\ref{app:prior-construction}).
For each (sample, span, layer) we measure three quantities: top-$k$ overlap between the two priors' top-$k$ sets, with $k$ matched to the model's routing top-$k$; the per-element $|\Delta b|$ at the 90th percentile of the \emph{effective} log-prior bias actually added to the router logits; and the fraction of clamp upper-bound activations.

The within-community shuffle changes $\sim$46\% of the prior's top-$k$ entries on math20k and $\sim$60\% on coding20k, with a 90th-percentile $|\Delta b|$ in the effective log-prior bias of $\sim$0.47 and $\sim$0.82 respectively.
Set against the empirical per-row bias standard deviation of $\sim$0.62 and $\sim$0.65 (Appendix~\ref{app:prior-construction}), the perturbation is on the order of one within-row standard deviation, that is, a change at the scale of the logits themselves rather than a nominal one.
Global shuffle changes $\sim$90\% of top-$k$ entries with a 90th-percentile $|\Delta b|$ of 1.5--1.6, saturating the intervention as a strong negative control should.
Clamp upper-bound activation is low in both domains (1--2\%), which rules out the reading that the shuffle is silently suppressed by the \texttt{clamp(p, eps/E, c/E)} step.
The deflationary objection, that the within-community shuffle does not actually change the prior, is therefore not supported on the prior side.

\begin{table*}[t]
  \centering
  \small
  \resizebox{\textwidth}{!}{%
  \begin{tabular}{llllllll}
    \toprule
    \textbf{Dataset} & \textbf{Method} & \textbf{Prior} & \textbf{Constraint} & \textbf{ID} & \textbf{OOD} & \textbf{(ID+OOD)/2} & \textbf{$\Delta$ vs.\ RPB} \\
    \midrule
    math20k & RPB (unshuffled reference) & base & soft & 45.77 & 19.53 & 32.65 & reference \\
    math20k & RPB-hard (unshuffled)$^\dagger$ & base & hard & 39.94 & 12.18 & 26.06 & hard constraint alone: $-6.59$ \\
    math20k & within-hard & within-shuffled & hard & 30.70 & 12.81 & 21.75 & hard constraint + within-scramble: $-10.90$ \\
    math20k & global-hard & global-shuffled & hard & 3.97 & 3.40 & 3.68 & hard constraint + global-scramble: $-28.97$ \\
    coding20k & RPB (unshuffled reference) & base & soft & 41.37 & 38.70 & 40.04 & reference \\
    coding20k & RPB-hard (unshuffled)$^\dagger$ & base & hard & 33.90 & 28.04 & 30.97 & hard constraint alone: $-9.07$ \\
    coding20k & within-hard & within-shuffled & hard & 23.14 & 25.73 & 24.44 & hard constraint + within-scramble: $-15.60$ \\
    coding20k & global-hard & global-shuffled & hard & 12.91 & 20.48 & 16.69 & hard constraint + global-scramble: $-23.35$ \\
    \bottomrule
  \end{tabular}
  }
  \caption{Decomposition of prior hardness and shuffle scope on Moonlight-16B-A3B. The RPB reference and the shuffled controls are three-seed means over the same nine-benchmark set as Table~\ref{tab:moonlight-summary}. $^\dagger$RPB-hard is single-seed (s42) from the $\beta$ sweep of Appendix~\ref{sec:appendix-beta-operating} and is reported for contrast only.}
  \label{tab:hardness-scope}
\end{table*}

\subsection{Hardness vs.\ Shuffle Scope Decomposition}
\label{app:hardness-scope}

The within-hard and global-hard rows of Table~\ref{tab:hardness-scope} separate two effects that are otherwise easy to confuse, the hardness of the constraint and the scope of label corruption.
The table reports the unweighted mean of the in-domain and out-of-domain figures, $(\mathrm{ID} + \mathrm{OOD})/2$, together with the $\Delta$ against the soft RPB reference computed from the same column, so every entry is recomputable from the columns shown.

Reference rows mix seed budgets, since the soft RPB reference and the shuffled controls are three-seed means (\S\ref{sec:results-headline}, \S\ref{sec:results-shuffled}) while RPB-hard is single-seed (Appendix~\ref{sec:appendix-beta-operating}).
Cross-protocol noise is on the order of a few tenths of a point and the gaps in the $\Delta$ column are at least an order of magnitude larger, so the qualitative ordering is unaffected.

\section{Reproducibility Protocol}
\label{app:repro}

\subsection{Compute Environment}
\label{app:compute}

All post-training runs execute on 16 nodes of 8 NVIDIA H200 GPUs, for a world size of 128, under Megatron-Bridge.
The parallel degrees are tensor 2, pipeline 1, context 1, expert 8, and expert-tensor 1, giving a data-parallel width of 64; sequence parallelism is enabled, and selective activation recomputation covers the layernorm, MoE, MoE-activation, and core-attention modules.
Training runs in mixed \texttt{bf16} on \texttt{torch} 2.10 with CUDA 12.9 and NCCL 2.27.
Evaluation runs separately under vLLM 0.10.2 on 8 GPUs per worker.

\subsection{Corpus Construction}
\label{app:corpus}

Both corpora are rendered from GLM-5.1-Reasoning-1M-Cleaned~\cite{glm51_reasoning_1m_cleaned} with the target model's own chat template, so row counts are identical across model families.
The \texttt{main} subset holds 328{,}033 rows; filtering it for fenced code blocks yields 199{,}704 candidate coding rows, and the \texttt{Math} topical subset holds 22{,}097 rows.
Each domain is then reduced to exactly 20{,}000 rows by a deterministic scorer rather than by random sampling: every row is scored once, the top 20{,}000 are kept, and they are written back in source order so that row identifiers remain stable.
For math20k this discards 2{,}097 rows, or 9.5\% of the topical subset.

The score sums a length term, a domain term, and a set of hard penalties; no model is used in the loop.
The length term scores prompt and answer separately against accepted bands --- 40 to 4{,}000 characters for prompts, centered at 500, and 200 to 16{,}000 for answers, centered at 2{,}500 --- with the answer term weighted more heavily, and rows outside a band are penalized in proportion to their distance from it.
The domain term rewards fenced code blocks and code keywords on coding20k and \LaTeX{} markup on math20k, together with numeric density, the presence of structural markers in the answer, and a minimum length of 80 words; each component is capped so that no single signal dominates.
The penalties remove rows that are empty, that retain a reasoning-channel tag from the source rendering, that lack the assistant turn marker, that contain repeated punctuation runs or degenerate phrases, or that exceed 32{,}000 characters in total.

\subsection{Prompting, Sampling and Answer Extraction}
\label{app:sampling}

Generation uses vLLM with temperature 0.6, nucleus sampling at $p{=}0.95$, and a generation budget of 8{,}192 tokens.
Each checkpoint is decoded five times from a fixed base sampling seed of 42, incremented per repeat, and the five scores are averaged to the avg@5 seed score of Appendix~\ref{app:seed-protocol}.
Answers are extracted deterministically in a fixed order: the last \verb|\boxed{}| span, located by a brace-balanced scan so that nested expressions survive; failing that, a \texttt{\#\#\#\#}-delimited final answer; failing that, the last number in the response.
For the multiple-choice benchmarks the option order is shuffled once with seed 42 and held fixed across all conditions, so every method sees the same permutation.

Prompts are zero-shot and are assembled from four task-type templates rather than per-benchmark ones, so that every condition sees an identical instruction for a given task family.
The \emph{math} template (GSM8K, MATH-500) pairs a system turn asking for step-by-step reasoning with a user turn that restates the problem and requires the final answer inside \verb|\boxed{}|.
The \emph{multiple-choice} template (MMLU, MMLU-STEM, MMLU-Pro, GPQA, GPQA-Diamond) asks for brief reasoning followed by a line of the form \texttt{Answer: X} naming the option letter.
The \emph{code} template (HumanEval, MBPP) asks for the complete solution in a single fenced \texttt{python} block; HumanEval supplies the function signature to complete, and MBPP supplies the task description together with the tests the function must pass.
The three LiveBench tasks carry their own answer-format instructions inside the question text, so they are sent as a bare user turn with no system preamble, matching the official LiveBench protocol.
All four are rendered through the target model's own chat template before decoding.

\subsection{Full ID/OOD Benchmark Assignments}
\label{app:benchmark-suite}

For math20k, OOD includes the code benchmarks (HumanEval, MBPP, LiveBench Coding) and the multi-domain block of general, STEM, professional, and reasoning QA (MMLU-STEM, MMLU, MMLU-Pro, GPQA, GPQA-Diamond, LiveBench Reasoning).
For coding20k, OOD is the symmetric complement, namely the math benchmarks (GSM8K, MATH-500, LiveBench Math) plus the same multi-domain block.

The three LiveBench tasks are drawn from the \texttt{livebench/math}, \texttt{livebench/reasoning}, and \texttt{livebench/coding} test splits as retrieved in May 2026, giving 368, 200, and 128 questions respectively, with the coding split carrying LiveBench release dates 2024-06-24 and 2024-07-26.
LiveBench replaces questions between releases, so scores are comparable only within a fixed snapshot, and every condition here was evaluated against the same one.

\paragraph{The nine-benchmark suite ($\mathrm{Overall}@9$).}
Aggregate scores on Moonlight-16B-A3B are reported over a nine-benchmark subset of the twelve, namely GPQA-Diamond, GSM8K, HumanEval, LiveBench Coding, LiveBench Math, LiveBench Reasoning, MATH-500, MBPP, and MMLU-Pro.
MMLU-STEM, MMLU, and GPQA are held out of the aggregate because they overlap heavily with MMLU-Pro and GPQA-Diamond and would repeatedly count the same knowledge axis, and they are still reported per benchmark in Table~\ref{tab:moonlight-benchmark-full}.
Every Moonlight-16B-A3B row labeled $\mathrm{Overall}@9$, \textbf{ID}, or \textbf{OOD} uses this nine-benchmark suite, with ID and OOD the split above restricted to it (math20k: 3 ID / 6 OOD, and coding20k symmetrically).
Two tables report twelve-benchmark aggregates instead.
The \textbf{Avg} column of Table~\ref{tab:moonlight-benchmark-full} is the twelve-benchmark mean, so it does not match the $\mathrm{Overall}@9$ values of Table~\ref{tab:anchoring-forms}; because that table lists all twelve benchmarks, averaging its nine retained columns reproduces them.
The \textbf{ID}, \textbf{OOD}, and \textbf{Avg} columns of Table~\ref{tab:deepseek-robustness} likewise cover all twelve, but its per-benchmark companion Table~\ref{tab:deepseek-benchmark-detail} lists only the nine, so those columns cannot be recomputed from it.

\subsection{Seed and Evaluation Aggregation Protocol}
\label{app:seed-protocol}

Each training seed (s42 / s43 / s44 as applicable) yields one fine-tuned checkpoint per (method, domain) cell.
Each checkpoint is evaluated by \texttt{avg@5} over five sampling seeds, giving the \emph{seed score}, and multi-seed rows report \texttt{mean$\pm$sd} across training seeds.
Seed budgets differ between the two kinds of measurement.
For \emph{downstream performance}, the main interventional rows (RPB, LBL, SFT), the shuffled controls and Router-L2 all use three seeds (s42 / s43 / s44), while single-reference rows (Base and the hard variants) use s42 only.
For \emph{routing-graph metrics}, which require a separate probing pass over a checkpoint, RPB, LBL and SFT use three probed seeds, the shuffled controls use two (s42 / s43), and Router-L2 and the remaining single-reference conditions use one (s42).
Rows that mix the two are marked in the relevant table.
The main contrasts in \S\ref{sec:results-shuffled} (within-soft against RPB) and Appendix~\ref{sec:appendix-beta-operating} (soft against hard) use within-protocol matching, so the comparisons that carry argumentative weight are between rows with matched seed budgets.
Per-row counts are repeated in each table's caption.

\subsection{Training Hyperparameters}
\label{app:hyperparameters}

All Moonlight-16B-A3B runs share the same backbone, optimizer, schedule, and batch configuration.
Each corpus is trained for five epochs at a global batch size of 128 with micro-batch 1 and a sequence length of 8{,}192 without packing, which is 782 optimizer steps over 20{,}000 rows.
The optimizer is Adam ($\beta_1{=}0.9$, $\beta_2{=}0.95$) with a cosine schedule from $10^{-5}$ to $10^{-6}$, 40 warmup steps, weight decay 0.1 and gradient clipping at 1.0.
Interventions differ only in their auxiliary objective or router-attached bias: SFT runs use the cross-entropy loss alone; LBL runs add the standard load-balancing penalty at coefficient $10^{-3}$, with $10^{-4}$ included as a robustness check; RPB runs inject the log-prior bias defined in \S\ref{sec:method-rpb} at the configured $\beta$ and disable both the load-balancing objective and the router bias-update rule, so the prior is the only router-side signal; and Router-L2 runs add the penalty defined in \S\ref{sec:method-l2} at $\lambda \in \{10^2, 10^3, 10^4\}$.
DeepSeek-V2-Lite uses the analogous configuration with the same fine-tuning split.

\subsection{Prior Construction (RPB)}
\label{app:prior-construction}

The RPB prior tensor \texttt{pi\_base[x,s,$\ell$,e]} is computed in a single pass over the fine-tuning split with the frozen base model.
For each sample, per-token softmaxed router probabilities at every MoE layer are aggregated over Q/A spans to a per-span scalar, yielding shape $N_{\mathrm{samples}} \times 2 \times L \times E$, which is $20000 \times 2 \times 26 \times 64$ for Moonlight-16B-A3B.
The runtime conversion to a zero-mean log-prior bias applies a per-expert clamp \texttt{clamp(p, eps/E, c/E)} (defaults $\epsilon = 0.05$, $c = 5.0$), renormalizes, takes the log, and subtracts the expert-axis mean, after clamping, so the bias has exact zero mean.
The empirical per-row bias standard deviation is 0.616 on math20k and 0.654 on coding20k, with a range of $\approx 3.03$, and clamp upper-bound saturation is rare, at 1--2\% of cells (Table~\ref{tab:shuffle-strength}).
$\beta$ controls the overall scale, and main RPB rows use $\beta{=}1.0$ unless noted.

\subsection{Probing Data Size}
\label{app:probe-size}

Co-activation graphs and probe statistics (sharpness, cross-domain divergence, top-$k$ overlap to base) are computed on a held-out probing pool of $\sim 5{,}000$ tokens per domain.
The same tokens are used across all methods to keep metrics comparable within a (model, domain) cell.
For bias-free probing (\S\ref{sec:method-graph}), the fine-tuned checkpoint is loaded without the RPB prior tensor or any training-time hook attached, and the probe pool is fed in inference mode.

\subsection{Detector Configuration}
\label{app:detector-config}

Layer-wise Louvain detection~\cite{Blondel2008FastUO} uses the standard modularity objective $Q$~\cite{Newman2006ModularityAC} with resolution parameter $\gamma = 1.0$ and a maximum of 100 iterations or convergence within $\Delta Q < 10^{-6}$.
The input to Louvain is the symmetric, non-negative, zero-diagonal co-activation matrix $W^{(\ell)}$ per layer, after the edge-fraction threshold of \S\ref{sec:method-graph} retains the top 10\% of edges by weight, with sweeps over 0.05, 0.15 and 0.20 in Appendix~\ref{app:sweeps}.
Edge weights are not row-normalized, since row-normalization changes the modularity geometry, so we use the raw co-occurrence frequencies.
Leiden~\cite{Traag2018FromLT} is used in the robustness sweep at otherwise-matched parameters.
Detector seeds 0, 1, 2 and 42 are run in the appendix, while the main-paper rows use seed 0.

\subsection{Community NMI Aggregation}
\label{app:nmi-aggregation}

Per-layer NMI is computed with the standard symmetric normalization, $2I(X;Y)/(H(X)+H(Y))$.
The scalar community NMI we report per checkpoint is the unweighted \emph{layer mean} across all MoE layers, and $Q$ and $\Delta Q$ follow the same aggregation.

\section{Full Result Tables}
\label{app:full-tables}

\begin{table*}[t]
  \centering
  \small
  \resizebox{\textwidth}{!}{%
  \begin{tabular}{ll|cccc|cc}
    \toprule
    \textbf{Dataset} & \textbf{Method}
    & \multicolumn{4}{c|}{\textbf{Routing-Graph Metrics}}
    & \multicolumn{2}{c}{\textbf{Performance}} \\
    \cmidrule(lr){3-6}\cmidrule(lr){7-8}
    & & \textbf{Comm. NMI} & \textbf{TVD} & \textbf{JSD} & \textbf{top-$k$} & \textbf{ID} & \textbf{OOD} \\
    \midrule
    \multirow{4}{*}{math20k} & SFT & \score{0.575}{0.004} & \score{0.320}{0.000} & \score{0.093}{0.000} & \score{0.469}{0.001} & \score{29.44}{0.32} & \score{15.65}{0.38} \\
    & LBL & \score{0.554}{0.006} & \score{0.306}{0.001} & \score{0.082}{0.000} & \score{0.420}{0.002} & \score{31.91}{0.10} & \score{14.97}{0.51} \\
    & RPB & \bestscore{0.631}{0.003} & \score{0.310}{0.001} & \score{0.089}{0.000} & \bestscore{0.485}{0.001} & \bestscore{45.77}{0.84} & \bestscore{19.53}{0.28} \\
    & Router-L2$^{\dagger}$ & 0.628 & 0.312 & 0.090 & 0.480 & \score{44.38}{0.21} & \score{19.34}{0.33} \\
    \midrule
    \multirow{4}{*}{coding20k} & SFT & \score{0.588}{0.001} & \score{0.315}{0.001} & \score{0.084}{0.001} & \score{0.482}{0.002} & \score{28.34}{1.36} & \score{35.83}{0.49} \\
    & LBL & \score{0.544}{0.003} & \score{0.317}{0.001} & \score{0.081}{0.000} & \score{0.391}{0.002} & \score{30.73}{0.13} & \score{35.56}{0.38} \\
    & RPB & \score{0.629}{0.001} & \score{0.327}{0.001} & \score{0.088}{0.000} & \score{0.481}{0.001} & \score{41.37}{0.30} & \bestscore{38.70}{0.11} \\
    & Router-L2$^{\dagger}$ & \textbf{0.632} & 0.321 & 0.086 & \textbf{0.486} & \bestscore{41.50}{0.52} & \score{38.28}{0.37} \\
    \bottomrule
  \end{tabular}
  }
  \caption{Routing-graph metrics and downstream performance for SFT, LBL, RPB, and Router-L2 ($\lambda{=}10^4$) on math20k and coding20k, all on Moonlight-16B-A3B. \textbf{ID} and \textbf{OOD} are seed-level means over the nine retained benchmarks (Appendix~\ref{app:benchmark-suite}). For SFT, LBL and RPB every column is a three-seed mean $\pm$ sample standard deviation (s42 / s43 / s44), with each performance seed score itself an avg@5 over five evaluation repeats. For Router-L2 ($^{\dagger}$) only the performance columns are three-seed means, since its routing-graph metrics come from a single probed checkpoint (s42) and are reported without a spread. Higher community NMI and top-$k$ overlap indicate retention of base co-activation structure, while TVD and JSD measure marginal-distribution shift. Bold marks the best mean within each dataset block on community NMI, top-$k$ overlap, ID, and OOD, and TVD and JSD are left unbolded because neither direction is unambiguously better.}
  \label{tab:retention-routerl2}
\end{table*}

\subsection{Routing-Graph Metrics}
\label{app:routing-graph-metrics}

Table~\ref{tab:retention-routerl2} reports the routing-state measurements behind the retention hierarchy of \S\ref{sec:results-hierarchy}, together with downstream performance, for SFT, LBL, RPB and Router-L2.
Community NMI separates the two soft anchors from SFT and LBL on both splits, while TVD and JSD place all four methods in the same range and top-$k$ overlap leaves SFT between the two anchors on coding20k.

\subsection{Per-Benchmark Main Interventions}
\label{app:per-benchmark}

Table~\ref{tab:moonlight-benchmark-full} breaks the Moonlight-16B-A3B headline results (\S\ref{sec:results-headline}) into individual benchmarks, adding the base checkpoint as a reference row, and groups the benchmarks as multi-domain, code, and math.
Table~\ref{tab:deepseek-benchmark-detail} gives the same breakdown for the four main interventions on DeepSeek-V2-Lite at a single training seed (s42).
The DeepSeek-V2-Lite table is included for completeness of the scope condition of \S\ref{sec:results-deepseek-main} rather than as an independent multi-seed claim.

\begin{sidewaystable*}
  \centering
  \footnotesize
  \setlength{\tabcolsep}{3.5pt}
  \resizebox{\textheight}{!}{%
  \begin{tabular}{ll|cccccc|ccc|ccc|c}
    \toprule
    \textbf{Method} & \textbf{Dataset}
    & \multicolumn{6}{c|}{\textbf{Multi-Domain}}
    & \multicolumn{3}{c|}{\textbf{Code}}
    & \multicolumn{3}{c|}{\textbf{Math}}
    & \textbf{Overall} \\
    \cmidrule(lr){3-8}\cmidrule(lr){9-11}\cmidrule(lr){12-14}\cmidrule(lr){15-15}
    & & \textbf{MMLU-STEM} & \textbf{MMLU} & \textbf{MMLU-Pro} & \textbf{GPQA} & \textbf{GPQA-Diamond} & \textbf{LiveBench Reasoning}
    & \textbf{HumanEval} & \textbf{MBPP} & \textbf{LiveBench Code}
    & \textbf{GSM8K} & \textbf{MATH-500} & \textbf{LiveBench Math} & \textbf{Avg} \\
    \midrule
    none & Base & \score{26.26}{0.33} & \score{26.43}{1.00} & \score{13.64}{0.76} & \score{24.55}{1.10} & \score{21.11}{2.84} & \score{4.00}{2.55} & \score{16.46}{7.07} & \score{14.08}{7.67} & \score{6.09}{4.07} & \score{44.56}{19.96} & \score{26.24}{10.08} & \score{9.76}{3.26} & \score{19.43}{3.79} \\
    \midrule
    SFT & \multirow{3}{*}{\rotatebox{90}{\makecell{math}}} & \score{15.80}{0.44} & \score{13.71}{0.53} & \score{14.26}{0.30} & \score{10.28}{0.52} & \score{9.33}{0.51} & \score{9.73}{1.50} & \score{24.72}{2.17} & \score{24.19}{0.58} & \score{11.67}{1.06} & \score{59.57}{0.47} & \score{19.51}{0.30} & \score{9.24}{0.61} & \score{18.50}{0.23} \\
    LBL & & \score{15.17}{1.38} & \score{12.27}{1.83} & \score{13.58}{0.33} & \score{9.54}{1.06} & \score{8.92}{1.13} & \score{8.93}{0.45} & \score{23.41}{1.06} & \score{23.29}{2.45} & \score{11.67}{1.40} & \score{63.85}{0.32} & \score{22.08}{0.45} & \score{9.79}{0.06} & \score{18.54}{0.58} \\
    RPB & & \bestscore{19.16}{0.75} & \bestscore{16.27}{0.95} & \bestscore{18.59}{0.79} & \bestscore{11.49}{0.84} & \bestscore{11.41}{1.11} & \bestscore{11.37}{0.46} & \bestscore{33.01}{1.83} & \bestscore{26.57}{4.20} & \bestscore{16.20}{3.36} & \bestscore{76.38}{0.63} & \bestscore{41.88}{1.33} & \bestscore{19.05}{1.37} & \bestscore{25.11}{0.03} \\
    \midrule
    SFT & \multirow{3}{*}{\rotatebox{90}{\makecell{code}}} & \score{56.51}{0.41} & \bestscore{56.31}{0.62} & \score{35.80}{0.42} & \score{28.17}{0.54} & \score{27.74}{1.47} & \bestscore{13.83}{1.78} & \score{43.58}{2.24} & \score{30.88}{2.03} & \score{10.57}{0.09} & \score{75.38}{0.42} & \score{43.29}{1.44} & \score{18.93}{0.64} & \score{36.75}{0.48} \\
    LBL & & \score{51.79}{1.43} & \score{49.91}{1.46} & \score{33.66}{0.27} & \score{27.31}{0.75} & \bestscore{28.69}{0.52} & \score{12.27}{1.68} & \score{46.46}{1.28} & \score{30.11}{0.88} & \score{15.62}{0.98} & \score{76.85}{0.33} & \score{42.84}{0.22} & \score{19.07}{0.24} & \score{36.21}{0.36} \\
    RPB & & \bestscore{57.89}{1.81} & \score{55.21}{2.70} & \bestscore{37.81}{1.03} & \bestscore{28.96}{0.57} & \score{27.88}{1.49} & \score{13.33}{0.60} & \bestscore{66.54}{0.74} & \bestscore{37.63}{1.10} & \bestscore{19.95}{0.39} & \bestscore{81.16}{0.47} & \bestscore{50.48}{0.33} & \bestscore{21.56}{0.43} & \bestscore{41.53}{0.35} \\
    \bottomrule
  \end{tabular}
  }
  \caption{Moonlight-16B-A3B per-benchmark results grouped by fine-tuning dataset and benchmark family. For the three fine-tuning methods, each cell is the mean $\pm$ sample standard deviation across three training seeds (s42 / s43 / s44), with each seed score itself an avg@5 over five evaluation repeats, matching the seed protocol of Table~\ref{tab:moonlight-summary}. The \textbf{Avg} column, unlike that table, is the mean over all twelve benchmarks, and averaging the nine retained columns instead reproduces the $\mathrm{Overall}@9$ values of Table~\ref{tab:anchoring-forms}. The \textbf{Base} row is the pretrained checkpoint before fine-tuning, which has no training seed, so its spread is over the five evaluation repeats. Bold indicates the best mean per benchmark column within each fine-tuning-dataset block.}
  \label{tab:moonlight-benchmark-full}

  \vspace{2.5em}

  \resizebox{\textheight}{!}{%
  \begin{tabular}{ll|ccc|ccc|ccc|c}
    \toprule
    \textbf{Dataset} & \textbf{Method}
    & \multicolumn{3}{c|}{\textbf{Multi-Domain}}
    & \multicolumn{3}{c|}{\textbf{Math}}
    & \multicolumn{3}{c|}{\textbf{Code}}
    & \textbf{Overall} \\
    \cmidrule(lr){3-5}\cmidrule(lr){6-8}\cmidrule(lr){9-11}\cmidrule(lr){12-12}
    & & \textbf{MMLU-Pro} & \textbf{GPQA-Diamond} & \textbf{LiveBench Reasoning}
    & \textbf{GSM8K} & \textbf{MATH-500} & \textbf{LiveBench Math}
    & \textbf{HumanEval} & \textbf{MBPP} & \textbf{LiveBench Code}
    & \textbf{Avg} \\
    \midrule
    none & Base
    & \score{13.96}{1.27} & \score{17.47}{5.43} & \score{2.20}{0.57}
    & \score{4.40}{0.78} & \score{3.92}{1.66} & \score{2.99}{0.76}
    & \score{22.07}{3.21} & \score{27.24}{5.06} & \score{1.72}{1.40}
    & 10.66 \\
    \midrule
    \multirow{3}{*}{math20k} & SFT
    & \score{14.29}{3.27} & \score{6.77}{2.44} & \score{7.50}{2.06}
    & \score{18.98}{2.25} & \score{11.92}{1.15} & \bestscore{9.76}{0.97}
    & \score{29.51}{4.73} & \score{29.48}{3.07} & \score{6.56}{2.51}
    & 14.98 \\
    & LBL
    & \score{15.25}{3.44} & \score{7.17}{2.87} & \bestscore{9.10}{1.39}
    & \score{19.26}{1.60} & \score{11.28}{1.05} & \score{9.74}{1.03}
    & \bestscore{32.07}{3.24} & \score{30.64}{3.03} & \bestscore{7.50}{1.42}
    & 15.78 \\
    & RPB
    & \bestscore{17.13}{3.05} & \bestscore{10.20}{4.76} & \score{7.90}{2.95}
    & \bestscore{20.94}{0.66} & \bestscore{12.08}{0.92} & \score{9.30}{1.40}
    & \score{30.37}{4.34} & \bestscore{32.28}{2.41} & \score{6.41}{1.28}
    & \textbf{16.29} \\
    \midrule
    \multirow{3}{*}{coding20k} & SFT
    & \score{19.53}{0.95} & \bestscore{22.83}{3.97} & \score{8.40}{2.97}
    & \bestscore{25.53}{1.14} & \bestscore{11.24}{1.09} & \score{8.14}{1.74}
    & \score{31.83}{3.00} & \score{26.44}{2.62} & \bestscore{7.50}{1.52}
    & 17.94 \\
    & LBL
    & \score{18.31}{0.73} & \score{21.72}{5.70} & \bestscore{10.20}{1.57}
    & \score{24.90}{1.01} & \score{11.16}{1.40} & \bestscore{8.40}{1.19}
    & \score{31.95}{3.16} & \score{27.08}{2.42} & \score{6.56}{1.62}
    & 17.81 \\
    & RPB
    & \bestscore{21.34}{0.75} & \score{21.82}{4.51} & \score{8.60}{1.64}
    & \score{22.90}{1.40} & \score{11.12}{1.14} & \score{7.54}{0.70}
    & \bestscore{32.93}{2.89} & \bestscore{28.20}{0.76} & \bestscore{7.50}{2.63}
    & \textbf{17.99} \\
    \bottomrule
  \end{tabular}
  }
  \caption{DeepSeek-V2-Lite per-benchmark results for the four main interventions (Base / SFT / LBL / RPB) on math20k and coding20k. Each cell is mean $\pm$ standard deviation over five evaluation repeats (avg@5) at a single training seed (s42). \textbf{Avg} aggregates by averaging the nine retained benchmark columns. Bold indicates the best mean within each fine-tuning dataset block per benchmark column, and ties on the displayed precision are both bolded.}
  \label{tab:deepseek-benchmark-detail}
\end{sidewaystable*}

\subsection{Per-Benchmark Shuffled-Prior Controls}
\label{app:per-benchmark-shuffle}

Table~\ref{tab:shuffle-summary} in \S\ref{sec:results-shuffled} reports the soft and hard dissociation at the aggregate level, as one $\mathrm{Overall}@9$ figure and one community NMI per condition.
The per-benchmark breakdown is given here together with the matched unshuffled RPB row, so the ordering can be checked column by column rather than only at the aggregate level.
On Moonlight-16B-A3B the ordering is RPB $\approx$ within-soft $\gg$ within-hard $>$ global-hard on both fine-tuning splits (Table~\ref{tab:moonlight-shuffle-appendix}), over the three training seeds used elsewhere in the paper.
On DeepSeek-V2-Lite the soft against hard gap reproduces, while global-hard sits slightly above within-hard on both splits (Table~\ref{tab:deepseek-shuffle-benchmark}), so the monotone dependence on shuffle scope is a Moonlight-16B-A3B result rather than a general one.
The DeepSeek-V2-Lite table is at a single training seed, s42 throughout except coding20k within-soft, which falls back to s43, and the spread reported per cell is the within-seed across-rep standard deviation (avg@5 with sample sd).
The RPB rows of both tables are the reference values used to read the soft against hard dissociation in \S\ref{sec:results-shuffled}, where within-soft sits within seed noise of RPB on Moonlight-16B-A3B while the hard variants fall away from it.

\begin{sidewaystable*}
  \centering
  \footnotesize
  \setlength{\tabcolsep}{3.5pt}
  \resizebox{\textheight}{!}{%
  \begin{tabular}{ll|ccc|ccc|ccc|c}
    \toprule
    \textbf{Dataset} & \textbf{Method}
    & \multicolumn{3}{c|}{\textbf{Multi-Domain}}
    & \multicolumn{3}{c|}{\textbf{Math}}
    & \multicolumn{3}{c|}{\textbf{Code}}
    & \textbf{Overall} \\
    \cmidrule(lr){3-5}\cmidrule(lr){6-8}\cmidrule(lr){9-11}\cmidrule(lr){12-12}
    & & \textbf{MMLU-Pro} & \textbf{GPQA-Diamond} & \textbf{LiveBench Reasoning}
    & \textbf{GSM8K} & \textbf{MATH-500} & \textbf{LiveBench Math}
    & \textbf{HumanEval} & \textbf{MBPP} & \textbf{LiveBench Code}
    & \textbf{Avg} \\
    \midrule
    \multirow{4}{*}{math20k} & RPB
    & \score{18.59}{0.79} & \score{11.41}{1.11} & \score{11.37}{0.46}
    & \score{76.38}{0.63} & \score{41.88}{1.33} & \score{19.05}{1.37}
    & \score{33.01}{1.83} & \score{26.57}{4.20} & \score{16.20}{3.36}
    & 28.27 \\
    & within-soft
    & \score{18.50}{1.52} & \score{11.11}{1.67} & \score{10.50}{0.44}
    & \score{75.75}{1.59} & \score{42.68}{0.62} & \score{19.53}{0.43}
    & \score{30.73}{0.68} & \score{19.99}{1.06} & \score{16.30}{1.58}
    & 27.23 \\
    & within-hard
    & \score{10.65}{0.43} & \score{3.50}{0.56} & \score{8.80}{0.44}
    & \score{53.97}{2.57} & \score{26.89}{2.17} & \score{11.23}{0.44}
    & \score{21.99}{9.45} & \score{26.12}{1.18} & \score{5.78}{2.34}
    & 18.77 \\
    & global-hard
    & \score{9.53}{0.37} & \score{3.50}{3.54} & \score{6.93}{0.67}
    & \score{3.36}{1.53} & \score{3.83}{0.12} & \score{4.73}{1.44}
    & \score{0.04}{0.07} & \score{0.16}{0.18} & \score{0.21}{0.09}
    & 3.59 \\
    \midrule
    \multirow{4}{*}{coding20k} & RPB
    & \score{37.81}{1.03} & \score{27.88}{1.49} & \score{13.33}{0.60}
    & \score{81.16}{0.47} & \score{50.48}{0.33} & \score{21.56}{0.43}
    & \score{66.54}{0.74} & \score{37.63}{1.10} & \score{19.95}{0.39}
    & 39.59 \\
    & within-soft
    & \score{35.43}{0.75} & \score{27.58}{0.44} & \score{11.90}{0.36}
    & \score{80.83}{0.53} & \score{49.65}{0.19} & \score{21.13}{0.88}
    & \score{63.98}{3.14} & \score{41.53}{3.76} & \score{18.65}{0.80}
    & 38.96 \\
    & within-hard
    & \score{18.79}{2.21} & \score{20.37}{1.02} & \score{11.63}{0.31}
    & \score{60.49}{1.65} & \score{30.00}{1.29} & \score{13.12}{0.78}
    & \score{34.39}{2.66} & \score{24.72}{2.02} & \score{10.31}{0.87}
    & 24.87 \\
    & global-hard
    & \score{16.08}{0.49} & \score{23.00}{0.65} & \score{10.23}{1.44}
    & \score{45.50}{3.95} & \score{20.01}{2.48} & \score{8.07}{0.85}
    & \score{18.66}{7.82} & \score{15.48}{5.38} & \score{4.58}{2.96}
    & 17.96 \\
    \bottomrule
  \end{tabular}
  }
  \caption{Moonlight-16B-A3B per-benchmark results for the soft-anchoring reference (RPB) and shuffled-prior controls (within-soft, within-hard, global-hard) on math20k and coding20k. Each cell is mean $\pm$ standard deviation across three training seeds (s42 / s43 / s44), where each seed score is itself avg@5 over five evaluation repeats. \textbf{Avg} aggregates by averaging the nine retained benchmark columns.}
  \label{tab:moonlight-shuffle-appendix}

  \vspace{2.5em}

  \resizebox{\textheight}{!}{%
  \begin{tabular}{ll|ccc|ccc|ccc|c}
    \toprule
    \textbf{Dataset} & \textbf{Method}
    & \multicolumn{3}{c|}{\textbf{Multi-Domain}}
    & \multicolumn{3}{c|}{\textbf{Math}}
    & \multicolumn{3}{c|}{\textbf{Code}}
    & \textbf{Overall} \\
    \cmidrule(lr){3-5}\cmidrule(lr){6-8}\cmidrule(lr){9-11}\cmidrule(lr){12-12}
    & & \textbf{MMLU-Pro} & \textbf{GPQA-Diamond} & \textbf{LiveBench Reasoning}
    & \textbf{GSM8K} & \textbf{MATH-500} & \textbf{LiveBench Math}
    & \textbf{HumanEval} & \textbf{MBPP} & \textbf{LiveBench Code}
    & \textbf{Avg} \\
    \midrule
    \multirow{4}{*}{math20k} & RPB
    & \score{17.13}{3.05} & \score{10.20}{4.76} & \score{7.90}{2.95}
    & \score{20.94}{0.66} & \score{12.08}{0.92} & \score{9.30}{1.40}
    & \score{30.37}{4.34} & \score{32.28}{2.41} & \score{6.41}{1.28}
    & 16.29 \\
    & within-soft
    & \score{16.30}{3.23} & \score{6.97}{4.39} & \score{9.10}{1.52}
    & \score{21.85}{1.35} & \score{12.00}{1.46} & \score{9.92}{1.91}
    & \score{27.32}{3.94} & \score{32.76}{1.64} & \score{6.88}{2.56}
    & 15.90 \\
    & within-hard
    & \score{10.15}{0.23} & \score{1.82}{0.77} & \score{6.00}{1.37}
    & \score{12.83}{1.32} & \score{7.64}{0.95} & \score{8.48}{1.40}
    & \score{0.98}{0.92} & \score{2.12}{0.73} & \score{0.16}{0.35}
    & 5.58 \\
    & global-hard
    & \score{10.15}{0.13} & \score{2.32}{1.05} & \score{7.40}{2.82}
    & \score{12.57}{1.27} & \score{7.20}{0.75} & \score{8.57}{1.05}
    & \score{7.20}{1.46} & \score{13.56}{1.53} & \score{1.25}{0.89}
    & 7.80 \\
    \midrule
    \multirow{4}{*}{coding20k} & RPB
    & \score{21.34}{0.75} & \score{21.82}{4.51} & \score{8.60}{1.64}
    & \score{22.90}{1.40} & \score{11.12}{1.14} & \score{7.54}{0.70}
    & \score{32.93}{2.89} & \score{28.20}{0.76} & \score{7.50}{2.63}
    & 17.99 \\
    & within-soft
    & \score{21.80}{0.42} & \score{22.22}{1.18} & \score{8.50}{2.18}
    & \score{23.28}{0.52} & \score{11.80}{1.02} & \score{7.89}{1.01}
    & \score{32.93}{3.48} & \score{29.36}{1.95} & \score{8.28}{2.45}
    & 18.45 \\
    & within-hard
    & \score{14.10}{0.42} & \score{20.20}{2.20} & \score{7.00}{1.00}
    & \score{9.16}{0.53} & \score{5.56}{0.99} & \score{5.89}{0.95}
    & \score{4.63}{1.02} & \score{7.56}{0.85} & \score{0.62}{0.35}
    & 8.30 \\
    & global-hard
    & \score{13.67}{0.53} & \score{21.92}{4.84} & \score{7.40}{1.29}
    & \score{10.52}{1.29} & \score{4.72}{1.43} & \score{5.32}{0.85}
    & \score{7.32}{3.14} & \score{9.04}{1.77} & \score{1.56}{1.10}
    & 9.05 \\
    \bottomrule
  \end{tabular}
  }
  \caption{DeepSeek-V2-Lite per-benchmark results for the soft-anchoring reference (RPB) and shuffled-prior controls (within-soft, within-hard, global-hard) on math20k and coding20k. Each cell is mean $\pm$ standard deviation over five evaluation repeats (avg@5) at a single training seed (s42 where available, and coding20k within-soft uses s43 because s42 was not trained). \textbf{Avg} aggregates by averaging the nine retained benchmark columns.}
  \label{tab:deepseek-shuffle-benchmark}
\end{sidewaystable*}

\end{document}